\documentclass[letterpaper, 10 pt, conference]{ieeeconf}

\IEEEoverridecommandlockouts  % This command is only needed if you want to use the \thanks command
\makeatletter
\def\endthebibliography{%
  \def\@noitemerr{\@latex@warning{Empty `thebibliography' environment}}%
  \endlist
}

\newcommand*\titleheader[1]{\gdef\@titleheader{#1}}
\AtBeginDocument{%
  \let\st@red@title\@title
  \def\@title{%
    \bgroup\normalfont\large\centering\@titleheader\par\egroup
    \vskip0.6em\st@red@title}
}
\makeatother

\titleheader{\normalsize To appear in \textit{Proceedings of International Conference on Robotics and Automation 2020}, Paris, France}

\usepackage{color}

\title{\LARGE \bf
    Non-Prehensile Manipulation in Clutter with Human-In-The-Loop
}

\author{Rafael Papallas and Mehmet R. Dogar% <-this % stops a space
    \thanks{This research has received funding from the European Union's Horizon 2020 research 
    and innovation programme under the Marie Sklodowska-Curie grants agreement 
    No. 746143, and from the UK Engineering and Physical Sciences Research Council 
    under grant EP/N509681/1, EP/P019560/1 and EP/R031193/1.}
\thanks{Authors are with the School of Computing, University of Leeds, United Kingdom
        {\tt\small \{r.papallas, m.r.dogar\}@leeds.ac.uk}}%
}

\usepackage{tabularx,booktabs}
\newcolumntype{C}{>{\centering\arraybackslash}X} % centered version of "X" type
\usepackage{hhline}
\usepackage{multirow}

\usepackage{caption}
\usepackage{subcaption}
\usepackage{graphicx,xcolor} 

\usepackage{algorithm}% http://ctan.org/pkg/algorithm
\usepackage[noend]{algpseudocode}% http://ctan.org/pkg/algorithmicx
\algnewcommand{\IIf}[1]{\State\algorithmicif\ #1\ \algorithmicthen}
\algnewcommand{\EndIIf}{\unskip\ \algorithmicend\ \algorithmicif}
\algdef{SE}[DOWHILE]{Do}{doWhile}{\algorithmicdo}[1]{\algorithmicwhile\ #1}%

\usepackage{mathtools}
\usepackage{bm}
\usepackage{esvect}
\usepackage{amssymb}
\usepackage{xspace}

\usepackage{scalerel}

\usepackage[english]{babel}

\usepackage[utf8]{inputenc}

\usepackage{csquotes}

\usepackage{graphicx,import}
\graphicspath{{./sections/introduction/images/},
              {./sections/conclusions/images/},
              {./sections/formulation/images/},
              {./sections/results/images/scenes/},
              {./sections/results/images/scenes_real_world/},
              {./sections/results/images/real_robot_experiment_1/},
              {./sections/results/images/real_robot_experiment_2/},
              {./sections/results/images/},
              {./sections/implementation/images/},
}

\usepackage{booktabs}

\usepackage[per-mode=symbol]{siunitx}

\usepackage{placeins}

\usepackage{hyperref}

\usepackage[capitalize]{cleveref}
\Crefname{algorithm}{Alg.}{Algs.}
\Crefname{section}{Sec.}{Secs}
\Crefname{line}{line}{lines}

\usepackage{microtype}
\usepackage[noadjust]{cite}
\newcommand{\acronymourapproach}{GRTC\xspace}

\newcommand{\acronymbaseline}{RTC\xspace}
\newcommand{\lreaching}{reaching through clutter\xspace}

\newcommand{\ourplanneracronym}{\acronymourapproach-HITL\xspace}
\newcommand{\heuristicplannername}{\acronymourapproach-Heuristic\xspace}
\newcommand{\ourplannerinfullacronym}{Guided-RTC with Human-In-The-Loop (\ourplanneracronym)\xspace}

\newcommand{\initialState}{q_{0}}
\newcommand{\goalStates}{Q_{goals}}
\newcommand{\goalState}{q_{n}}
\newcommand{\stateSpace}{Q}
\newcommand{\controlSpace}{U}

\newcommand{\systemDynamics}{f}
\newcommand{\systemDynamicsFunction}{{\systemDynamics: \stateSpace \times \controlSpace \to \stateSpace}}

\newcommand{\object}[1]{o_{#1}}
\newcommand{\guideState}[1]{{(\object{#1}, x_{#1}, y_{#1})}}

\begin{document}
    % Repeatable typeset (definitions etc) are placed here.
    \newcommand{\testing}{Feedback and Revaluation}

    \maketitle
    \thispagestyle{empty}
    \pagestyle{empty}

    \begin{abstract}
We propose a human-operator guided planning approach to pushing-based
manipulation in clutter. 
Most recent approaches to manipulation in clutter employs
randomized planning. The problem, however, remains a challenging one where the
planning times are still in the order of tens of seconds or minutes, and the
success rates are low for difficult instances of the problem. We build on these
control-based randomized planning approaches, but we investigate using them in conjunction
with human-operator input. In our framework, the human operator supplies a
high-level plan, in the form of an ordered sequence of objects and their approximate goal positions.
We present experiments in simulation and on a real robotic setup, where we compare the success rate and
planning times of our human-in-the-loop approach with fully autonomous sampling-based planners.
We show that with a minimal amount of human input, the low-level planner can solve the
problem faster and with higher success rates.
\end{abstract}

    \section{Introduction}
    \label{sec:introduction}

\begin{figure}[!t]
        \captionsetup[subfigure]{aboveskip=1.0pt,belowskip=1.0pt}
        \centering
        \begin{subfigure}{.32\linewidth}
            \centering
            \tiny
            \def\svgwidth{0.95\columnwidth}
%% Accompanies image file '1.pdf' (pdf, eps, ps)
\begingroup%
  \makeatletter%
  \providecommand\color[2][]{%
    \errmessage{(Inkscape) Color is used for the text in Inkscape, but the package 'color.sty' is not loaded}%
    \renewcommand\color[2][]{}%
  }%
  \providecommand\transparent[1]{%
    \errmessage{(Inkscape) Transparency is used (non-zero) for the text in Inkscape, but the package 'transparent.sty' is not loaded}%
    \renewcommand\transparent[1]{}%
  }%
  \providecommand\rotatebox[2]{#2}%
  \newcommand*\fsize{\dimexpr\f@size pt\relax}%
  \newcommand*\lineheight[1]{\fontsize{\fsize}{#1\fsize}\selectfont}%
  \ifx\svgwidth\undefined%
    \setlength{\unitlength}{950.00006104bp}%
    \ifx\svgscale\undefined%
      \relax%
    \else%
      \setlength{\unitlength}{\unitlength * \real{\svgscale}}%
    \fi%
  \else%
    \setlength{\unitlength}{\svgwidth}%
  \fi%
  \global\let\svgwidth\undefined%
  \global\let\svgscale\undefined%
  \makeatother%
  \begin{picture}(1,0.94736836)%
    \lineheight{1}%
    \setlength\tabcolsep{0pt}%
    \put(0,0){\includegraphics[width=\unitlength,page=1]{1.pdf}}%
    \put(0.45812889,0.57002671){\color[rgb]{0,0,0}\makebox(0,0)[lt]{\lineheight{1.25}\smash{\begin{tabular}[t]{l}$o_2$\end{tabular}}}}%
    \put(0.7494225,0.653492){\color[rgb]{0,0,0}\makebox(0,0)[lt]{\lineheight{1.25}\smash{\begin{tabular}[t]{l}$o_g$\end{tabular}}}}%
  \end{picture}%
\endgroup%

            \caption{}
            \label{fig:guidance_explained_1}
        \end{subfigure}%
        \begin{subfigure}{.32\linewidth}    
            \centering
            \tiny
            \def\svgwidth{0.95\columnwidth}
            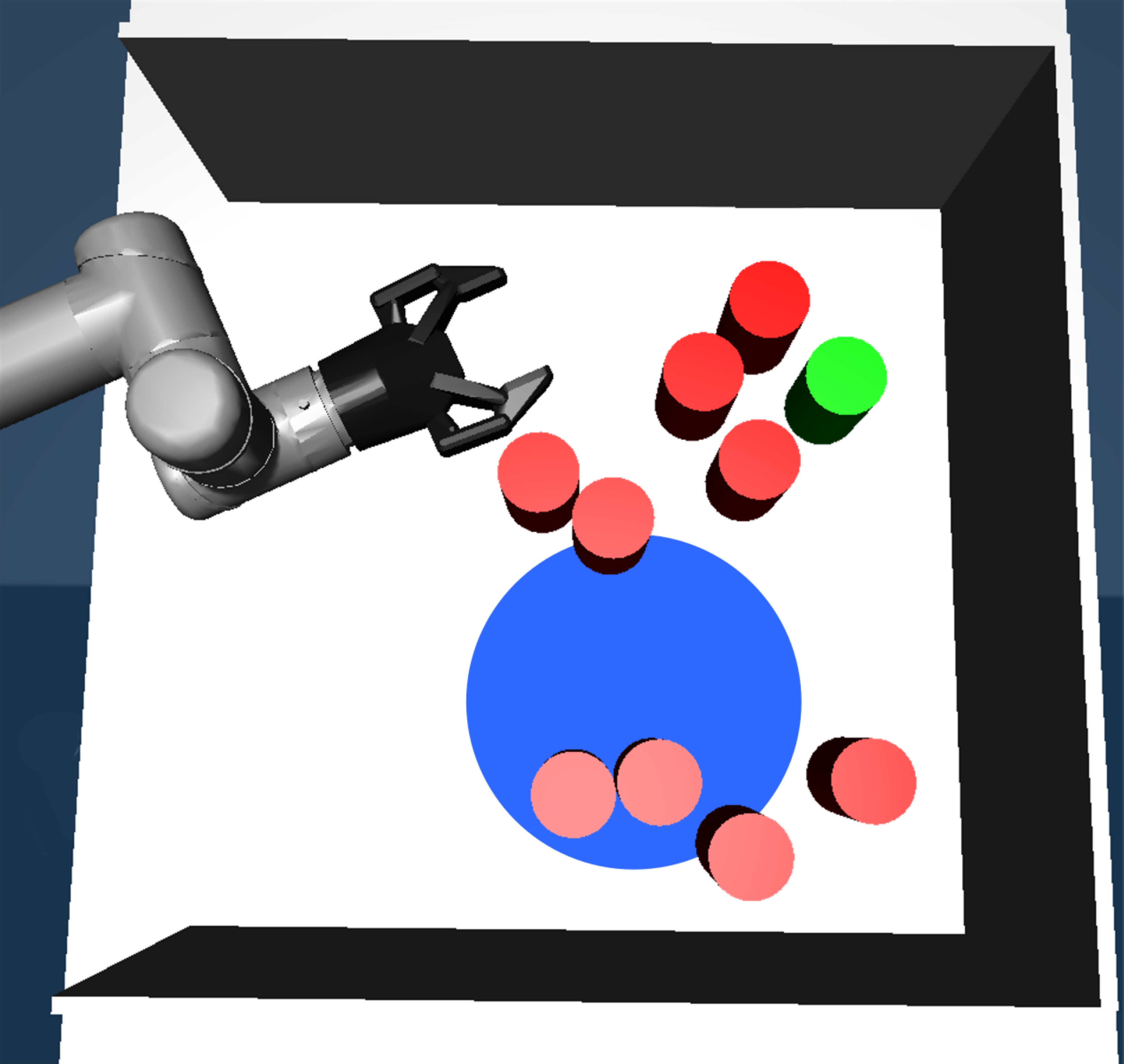
            \caption{}
            \label{fig:guidance_explained_2}
        \end{subfigure}%
        \begin{subfigure}{.32\linewidth}    
            \centering
            \tiny
            \def\svgwidth{0.95\columnwidth}
            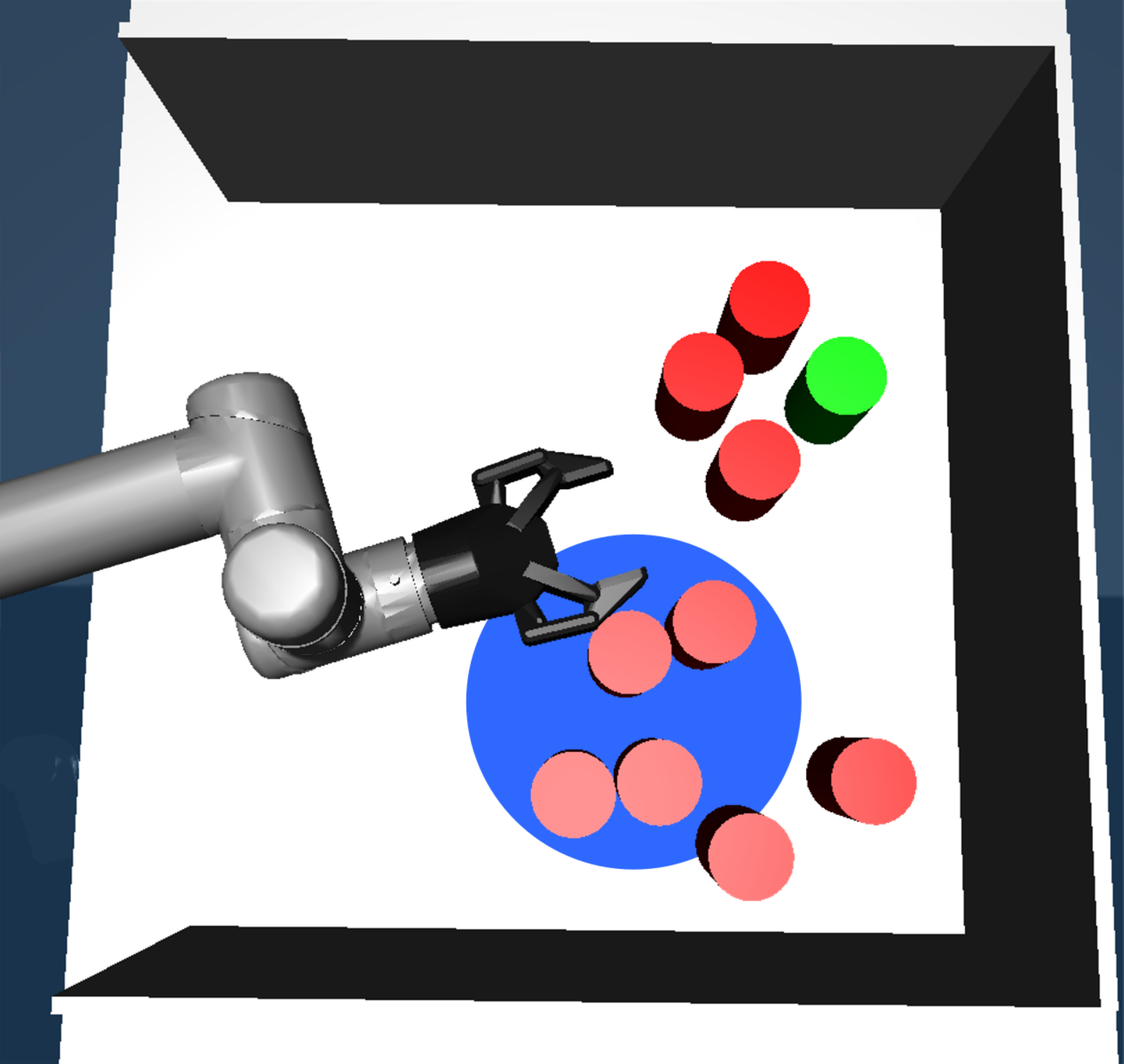
            \caption{}
            \label{fig:guidance_explained_3}
        \end{subfigure}

        \begin{subfigure}{.32\linewidth}    
            \centering
            \tiny
            \def\svgwidth{0.95\columnwidth}
            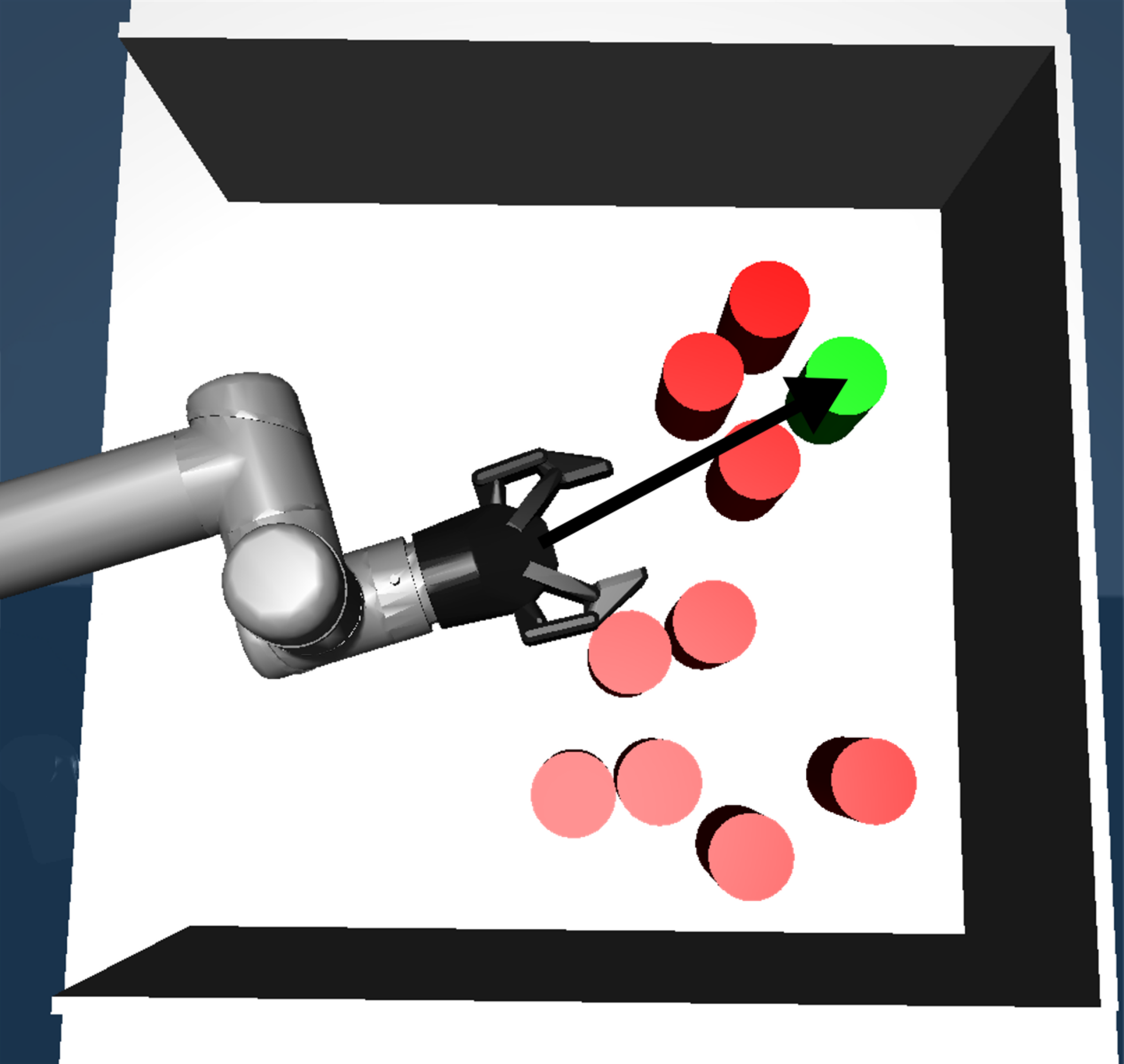
            \caption{}
            \label{fig:guidance_explained_4}
        \end{subfigure}%
        \begin{subfigure}{.32\linewidth}    
            \centering
            \tiny
            \def\svgwidth{0.95\columnwidth}
            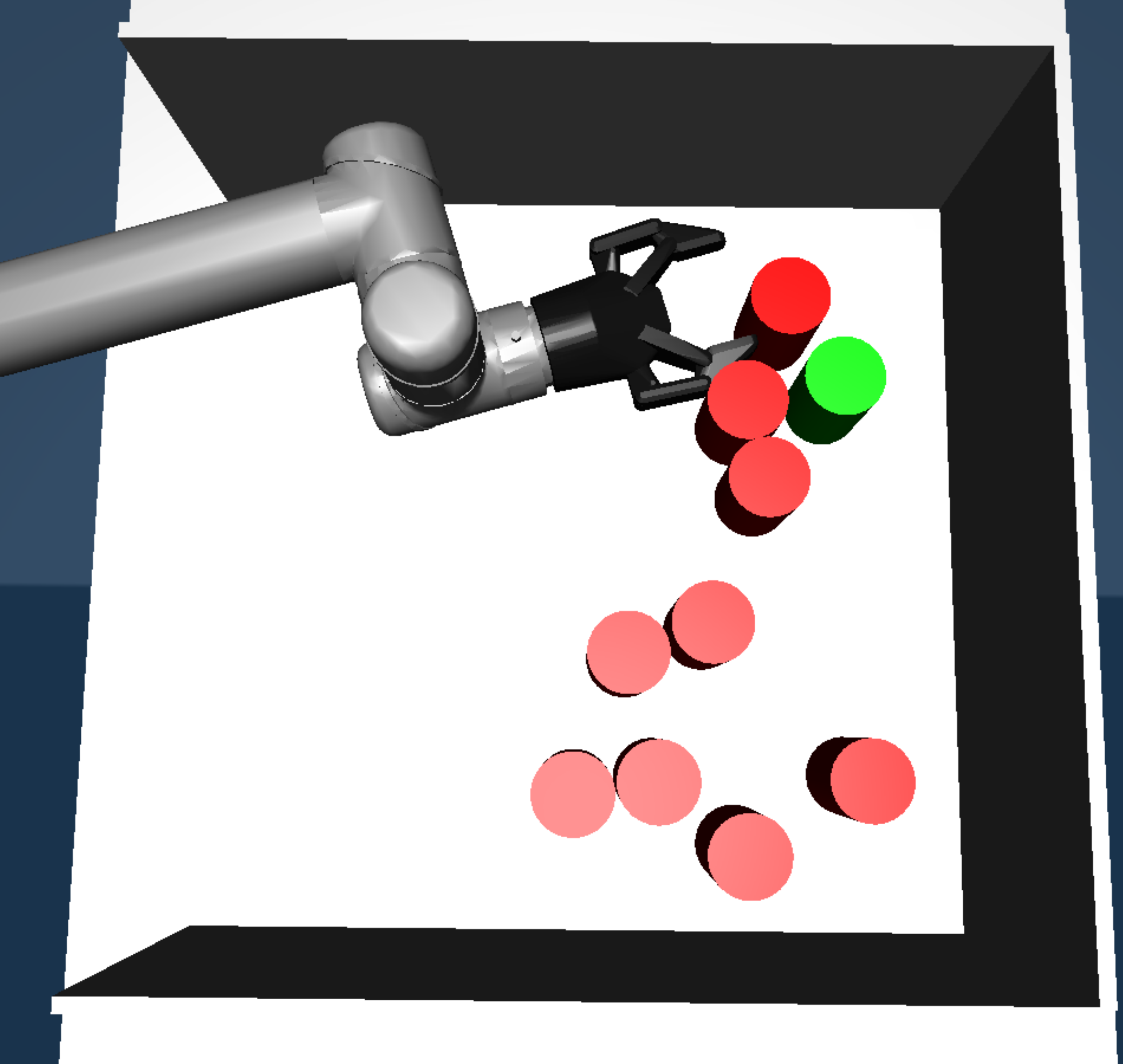
            \caption{}
            \label{fig:guidance_explained_5}
        \end{subfigure}%
        \begin{subfigure}{.325\linewidth}    
            \centering
            \tiny
            \def\svgwidth{0.95\columnwidth}
            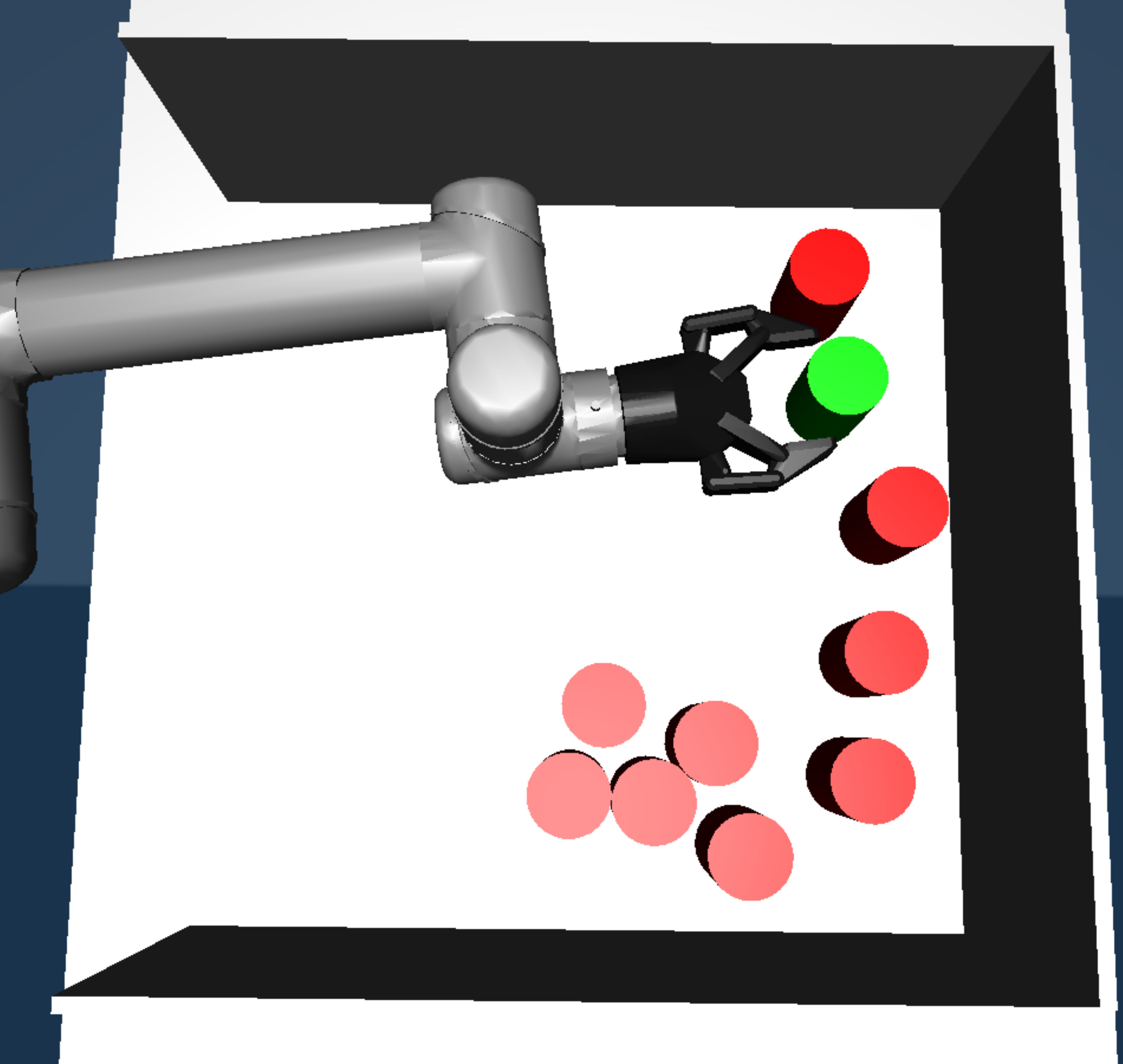
            \caption{}
            \label{fig:guidance_explained_6}
        \end{subfigure}%
        \caption{A human-operator guiding a robot to reach for the green goal
        object, $o_g$. Arrows indicate human interaction with the robot. 
        In (a) the operator indicates $o_2$ to be pushed to
        the blue target region. From (a) to (c) the robot plans to
        perform this push. In (d) the operator indicates to the robot
        to reach for the goal object. From (d) to (f) the robot plans to reach
        the goal object.}
        \label{fig:guidance_explained}
    \end{figure}

\begin{figure}[!b]
        \captionsetup[subfigure]{aboveskip=1.0pt,belowskip=1.0pt}
        \centering
        \begin{subfigure}{.25\linewidth}
            \centering
            \def\svgwidth{0.95\columnwidth}
            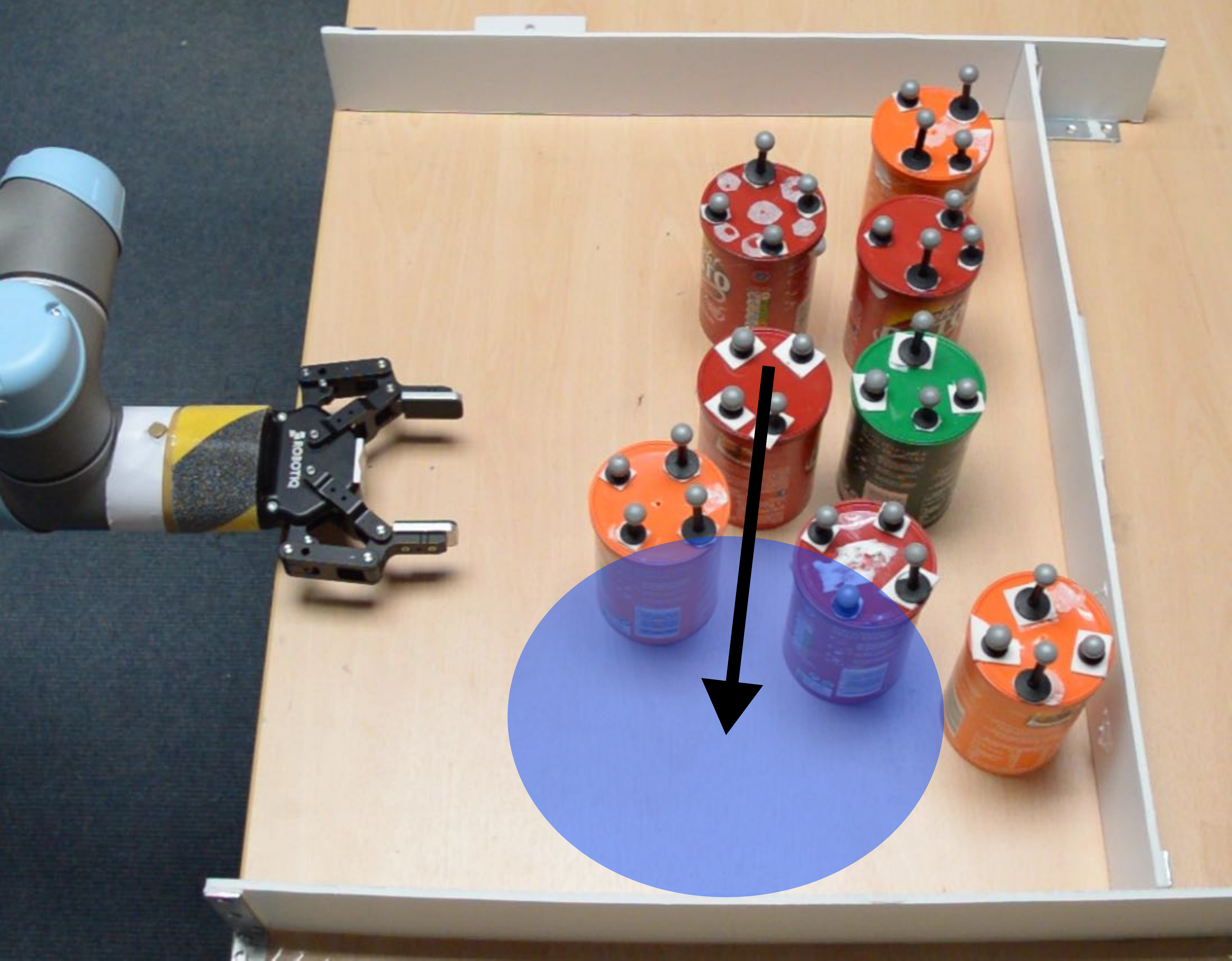
            \caption{}
            \label{fig:real_experiment_explained_1_1}
        \end{subfigure}%
        \begin{subfigure}{.25\linewidth}
            \centering
            \def\svgwidth{0.95\columnwidth}
            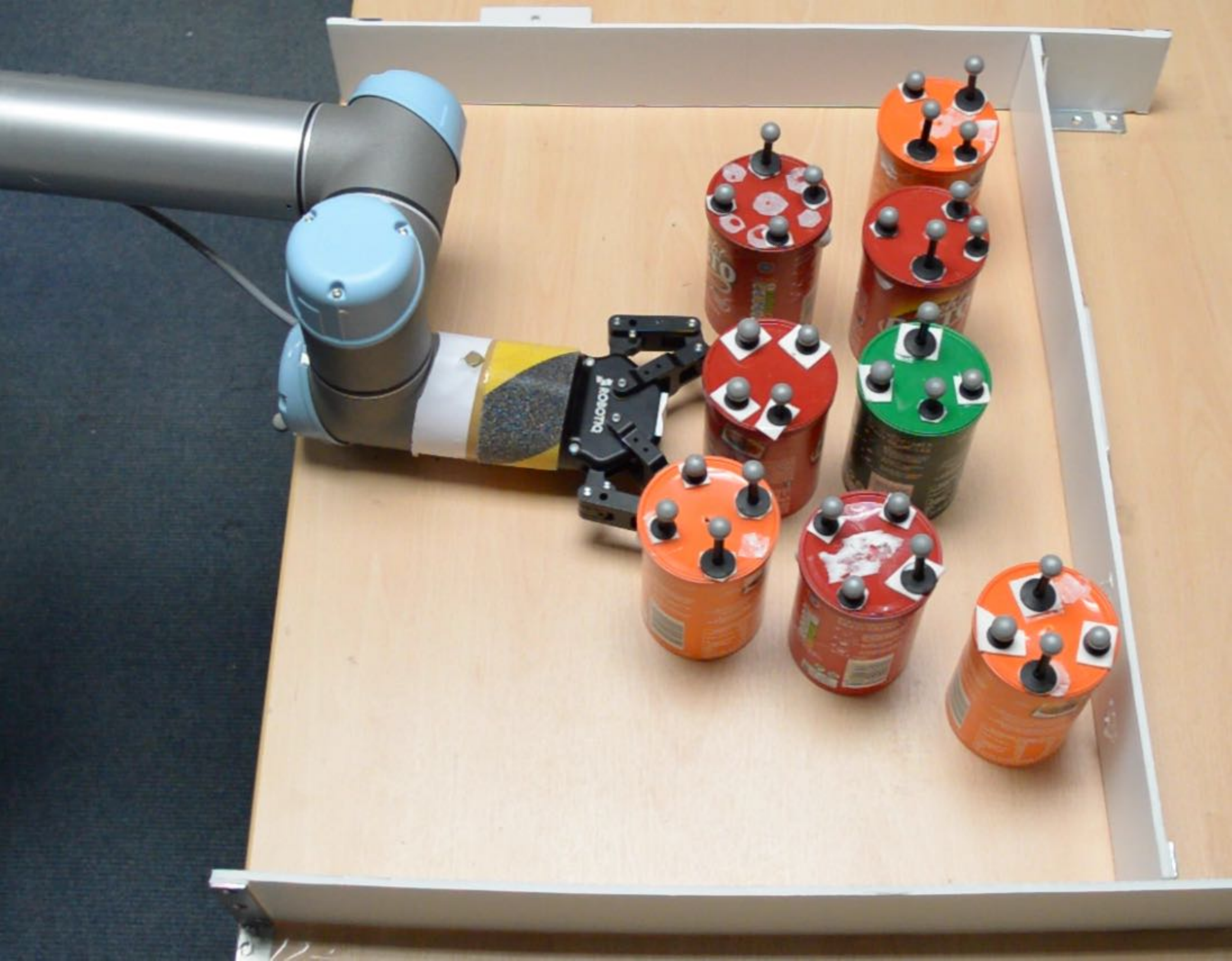
            \caption{}
            \label{fig:real_experiment_explained_1_2}
        \end{subfigure}%
        \begin{subfigure}{.25\linewidth}
            \centering
            \def\svgwidth{0.95\columnwidth}
            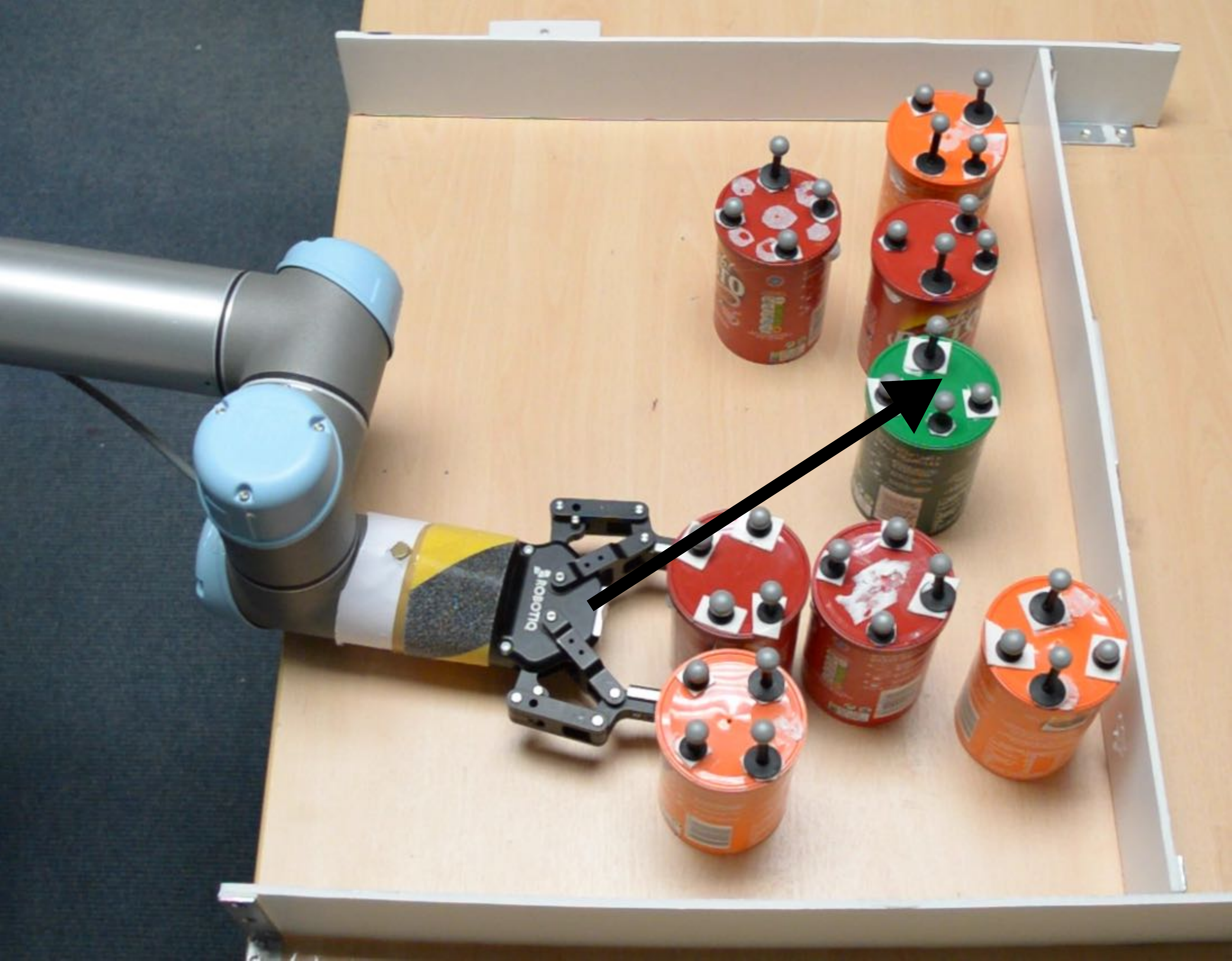
            \caption{}
            \label{fig:real_experiment_explained_1_3}
        \end{subfigure}%
        \begin{subfigure}{.25\linewidth}
            \centering
            \def\svgwidth{0.95\columnwidth}
            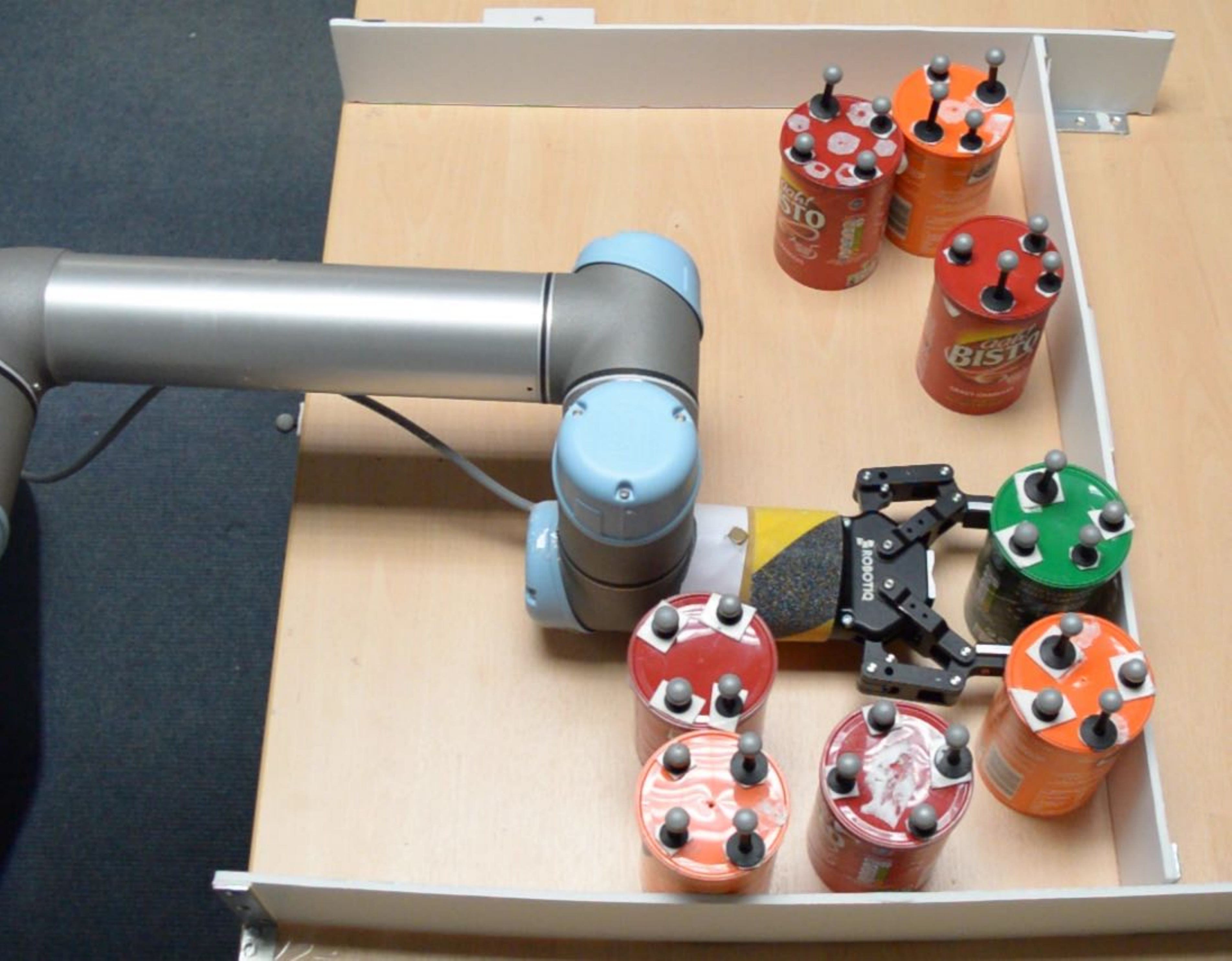
            \caption{}
            \label{fig:real_experiment_explained_1_4}
        \end{subfigure}
        \caption{Human-operator guiding a robot in the real-world.}
        \label{fig:real_experiment_explained_1}
    \end{figure}

    We propose a human-operator guided planning approach to pushing-based
    manipulation in clutter. We present example problems in 
    \cref{fig:guidance_explained,fig:real_experiment_explained_1}. The target
    of the robot is to reach and grasp the green object. To do this,
    however, the robot first has to push other objects out of
    the way (\cref{fig:guidance_explained_2} to
    \cref{fig:guidance_explained_5}). This requires the robot to 
    plan which objects to contact, where and how to push those objects so
    that it can reach the goal object. We present an approach
    to this problem where a human-in-the-loop provides a \textit{high-level
    plan}, which is used by a low-level planner to solve the problem.

    These \textit{\lreaching} problems are difficult to solve due
    to several reasons: First, the number of objects make the state space of
    high-dimensionality because the planner needs to reason about the robot state
    and all the movable objects. Second, this is an underactuated problem, since the
    objects cannot be controlled by the robot directly. Third,
    predicting the evolution of the system state requires running computationally
    expensive physics simulators, to predict how objects would move as a result of
    the robot pushing. Effective algorithms have been developed
    \cite{dogar2012physics,havur2014geometric,kitaev2015physics,
    haustein2015kinodynamic,moll2018randomized,bejjani2018,Nam2019,
    king2016rearrangement,bejjani2019learning,agboh2018real,huang2019large,
    kim2019retrieving},
    however, the problem remains a challenging one, where the planning times
    are still in the order of tens of seconds or minutes, and the success rates are
    low for difficult problems.

    Further study of the \lreaching problem is important to develop
    approaches to solve the problem more successfully and faster. It
    is a problem that has a potential for major near-term impact in warehouse
    robotics (where robots need to reach into shelves to retrieve objects) and
    personal home robots (where a robot might need to reach into a fridge or shelf
    to retrieve an object).  The Amazon Picking Challenge
    \cite{eppner2016lessons} was a competition which gained particular attention
    for this potential near-term impact of robotic manipulation to warehouse
    robotics. The algorithms that we currently have, however, are not able to solve
    \lreaching problems in the real world in a fast and consistent way.
    Here, we ask the question of whether human-operators can be used to provide
    a minimal amount of input that results in a significantly higher success
    rate and faster planning times.

    Most recent approaches to the \lreaching problem employs the power of randomized 
    \textit{kinodynamic planning}. Haustein et al.
    \cite{haustein2015kinodynamic} use a kinodynamic RRT
    \cite{lavalle1998rrt,lavalle2001randomized} planner to sample and generate a
    sequence of robot pushes on objects to reach a goal state. Muhayyuddin et al.
    \cite{moll2018randomized} use the KPIECE algorithm \cite{csucan2009kpiece} to
    solve this problem. These planners report some of the best performance (in terms of
    planning times and success rates) in this domain so far.

    We build on these kinodynamic planning approaches,
    but we investigate using them in conjunction with human-operator input. In our
    framework, the human operator supplies a \textit{high-level plan} to make the
    underlying planners solve the problem faster and with higher success rates.

    For example in \cref{fig:guidance_explained_1}, the human operator supplies the high-level action 
    of first pushing the object $o_2$ to the blue region. 
    A key point here is that pushing $o_2$ in \cref{fig:guidance_explained_1} 
    to its target region is itself a problem which requires kinodynamic planning through clutter, 
    since the object and the robot may need to contact and push other 
    objects during the motion. 
    %Therefore the action in \cref{fig:guidance_explained_1} requires the 
    %use of a planner that can potentially solve the original larger problem of 
    %reaching the green object through clutter. However, solving the original 
    %problem in highly cluttered settings can take a long infeasible amount 
    %of time. We propose that the human can suggest high-level actions that 
    %decompose the original problem into a sequence of easier problems of the 
    %same type. 
    The robot uses a kinodynamic planner to approach $o_2$ in
    \cref{fig:guidance_explained_2} and to push it to
    the target region in \cref{fig:guidance_explained_3}. 
    %The human operator's input here is 
    %limited to selection of the object and an approximate
    %goal location for that object. The actual pushing actions are planned by the
    %autonomous low-level planner. 
    In \cref{fig:guidance_explained_4} the operator 
    points directly the actual goal object (green). The kinodynamic planner 
    in \cref{fig:guidance_explained_5} finds a way to push
    other objects out of the way and 
    in \cref{fig:guidance_explained_6} successfully reaches the goal object.
    The human operator's role in the system is not to guide the robot all the
    way to the goal, but to provide key high-level actions to help
    the robot. %to reach for the goal object.
    At any point during the interaction, even at the very beginning, the
    operator can decide not to provide any further high-level actions
    (either because the scene is easy enough for the low-level
    planner or because the operator is busy) and she can command the system to
    plan directly for the actual goal object. The system degrades nicely to
    state-of-the-art kinodynamic planning if no high-level actions are
    provided.

    The use of high-level plans is related to recent work in robotic 
    hierarchical planning \cite{stilman2007manipulation,kaelbling2010hierarchical,dogar2011framework,havur2014geometric,lee2015hierarchical} and
    task-and-motion planning (TAMP) \cite{lagriffoul2014efficiently,wells2019learning,lee2019efficient}.
    This line of work shows that with a good high-level plan for a task, the search
    of the low-level motion planner can be guided to a relevant but restricted part
    of the search space, making the planner faster and more successful.
    Particularly relevant to our problem is the work from Stilman et al.  \cite{stilman2007manipulation}, which formulates the
    problem of \textit{manipulation/navigation among movable obstacles (NAMO)} as a
    high-level search over the orderings of objects to be moved, combined with a
    low-level motion planner that pick objects up and move in that order.
    We use a similar high-level plan structure, i.e. an ordering of objects, but we focus on
    non-prehensile manipulation of objects, rather than pick-and-place. 
    %In hierarchical approaches to non-prehensile manipulation, determining
    %which object to contact and where is usually the high-level problem. For
    %example, Lee et al. \cite{lee2015hierarchical} proposes hierarchical planning for
    %non-prehensile manipulation of a single object, where the high-level plan
    %includes the sequence of contacts to be made with the object. Similarly, in
    %our work the high-level decision is about the sequence of objects to
    %contact. 

    The hierarchical/TAMP planners above generate high-level plans
    autonomously.  Motivated by existing work in human-in-the-loop planning
    \cite{hwang1997human, bayazit2001enhancing, taix2012human, denny2016theory,
    islam2017online, denny2018general}, in this work we investigate the
    potential of using a human operator to suggest high-level plans. 
    The existing work in human-in-the-loop planning focuses on path planning and providing clues to a
    planner to guide it through the collision-free space.
    We explore a similar approach, but in the context of \textit{non-prehensile 
    pushing-based planning}, where human physical intuition can be useful.
    %In the domain of collision-free path planning 
    %The idea of using human input during robotic planning has been studied before.
    %There were a number of contributions on this topic
    %They introduce the idea of collaborative planning between a robot and a
    %human. 
    %In our framework we leverage human's intuition to choose the
    %appropriate action at the time to rearrange the environment to reach for a
    %goal object. 
    Other human-in-the-loop systems have been investigated
    for pick-and-place tasks \cite{leeper2012strategies,
    muszynski2012adjustable, witzig2013context, ciocarlie2012mobile} but to the 
    best of our knowledge, a human-in-the-loop approach has not been applied
    to non-prehensile physics-based manipulation before.

    We compare our method to using kinodynamic methods without any high-level
    plans, e.g. KPIECE and RRT. We also compare our method to hierarchical
    methods which generate high-level plans autonomously.  For the latter, we
    implemented a non-prehensile variation of the NAMO planner
    as well as an approach which uses a straight-line motion heuristic to generate candidate objects for the
    high-level plan.  We performed experiments in simulation and on a real
    robot, which shows that the human-in-the-loop approach produces more
    successful plans and faster planning times. This gain, of course, comes at
    the expense of a human operator's time. We show that this time is minimal
    and to evaluate this further, we experiment with a single human operator
    providing high-level plans in-parallel to multiple robots and present an
    analysis. We discuss whether such an approach may be feasible in a
    warehouse automation setting.  To support reproducibility, we provide the
    source code of all our algorithms and experiments in an open repository\footnote{\url{https://github.com/rpapallas/hitl_clutter}}.

%        While a good high-level plan can make the low-level planning problem easier to solve, the
%    autonomous generation of a good high-level plan is itself a computationally
%    expensive problem.  
%    For example, for our problem, i.e. \lreaching, a high-level planner would need
%    to search in a space of all possible permutations of objects, combined with all
%    possible goal locations for these objects. Furthermore, the high-level planner
%    must be able to choose high-level actions that are feasible for the low-level
%    planner. In our problem, this feasibility would mean checking/predicting whether the
%    robot would be able to push a certain object to a certain location, which
%    either requires the use of computationally expensive physics simulations, or a
%    heuristic to estimate the probability of successful push.
%
%    While these decisions are computationally expensive for an autonomous planner,
%    they can be easy for a human.  
%
%    \rafael{Talk about how we compare our work with RTC and Heuristic planners.}
%    \rafael{Provide a summary of the results section.}

    \section{Problem Formulation}%
    \label{sec:formulation}
    Our environment is comprised of a robot $r$, a set of movable obstacles $O$, and
    other static obstacles. The robot is allowed to interact with the movable obstacles, but not with the static ones.
    We also have ${o_g \in O}$ which is the \textit{goal} object to reach.

    We are interested in problems where the robot needs to reach for an 
    object in a cluttered shelf that is constrained from the top, and
    therefore we constrain the robot motion to the plane
    and its configuration space, $Q^r$, to $SE(2)$. The configuration of a movable 
    object ${i \in \{1, \dots, |O|\}}$, $q^i$, is its pose on the plane
    ($x, y, \theta$). We denote its configuration space as $Q^{i}$.
    The configuration space of the complete system is the Cartesian product
    ${Q = Q^r \times Q^{g} \times Q^{1} \times \dots \times Q^{|O|-1}}$.

    Let ${\initialState \in \stateSpace}$ be the initial
    configuration of the system, and ${\goalStates \subset \stateSpace}$ a set of possible
    goal configurations. A goal configuration, ${\goalState \in Q_{goals}}$, is defined as a configuration where $o_g$ is within
    the robot's end-effector (see \cref{fig:guidance_explained_6}).

    Let $\controlSpace$ be the control space comprised of the robot
    velocities. Let the system dynamics be defined as $\systemDynamicsFunction$ that propagates
    the system from ${q_t \in \stateSpace}$ with a control
    ${u_t \in \controlSpace}$. 
    
    We define the \textit{Reaching Through Clutter} (\acronymbaseline) problem as the tuple
    ${(\stateSpace, \controlSpace, \initialState, \goalStates, \systemDynamics)}$. The solution 
    to the problem is a sequence of
    controls from $\controlSpace$ that move
    the robot from $\initialState$ to a ${\goalState \in Q_{goals}}$.

    \section{Sampling-based Kinodynamic Planners}%
    \label{sec:rtc_planners}

    Two well known sampling-based kinodynamic planners are Rapidly-exploring Random 
    Trees (RRT) \cite{lavalle1998rrt, lavalle2001randomized} and
    Kinodynamic Motion Planning by Interior-Exterior Cell Exploration (KPIECE)
    \cite{csucan2009kpiece}. Both RRT and KPIECE have been used before 
    in the literature to solve problems similar to the \acronymbaseline problem
    \cite{haustein2015kinodynamic,moll2018randomized,rusu2009real,nieuwenhuisen2013mobile}.
    We use kinodynamic RRT and KPIECE in our work in two different ways: (1) as
    baseline RTC planners to compare against, and (2) as the low-level planners
    for the Guided-RTC Framework that accepts high-level actions (explained in \cref{sec:guidance_planner}).

    \noindent\textbf{Kinodynamic RRT}:
        RRT is a sampling-based planner that builds a
        tree from the initial state, $q_{0} \in Q$, and then samples a
        random state $q_{rand} \in Q$ and tries to extend the tree from the nearest
        neighbor $q_{near} \in Q$. Kinodynamic RRT samples controls to bring $q_{near}$
        near to $q_{rand}$. 

    \noindent\textbf{KPIECE}:
        KPIECE is a sampling-based planner that operates well in
        problems with complex dynamics \cite{csucan2009kpiece}. KPIECE builds a 
        tree from the state and control space until a goal is reached. KPIECE
        uses the notion of space coverage to guide its exploration in the state
        space by constructing a discretization of the state space. KPIECE
        samples a control from $\controlSpace$ and it tries to expand the tree
        and update the discretization. 

        \vspace{0.5em}

        In this work, when we plan with a kinodynamic planner (either RRT or KPIECE) we will
        use the notation \textit{kinodynamicPlanning($q_{start}$, goal)} with a
        start configuration of the system, $q_{start}$, and some \textit{goal} input.
        %To solve a \acronymbaseline problem, we will use \textit{kinodynamicPlanning($q_c$, $\goalStates$)}.

\section{Guided-RTC Framework}%
    \label{sec:guidance_planner}

    In this section, we describe a \textit{guided} system to solve RTC problems.
    A Guided-RTC system accepts high-level actions.
    A high-level action can suggest to push a particular
    obstacle object into a certain region, or it may suggest to reach for the goal
    object. We formally define a high-level action with the triple $\guideState{i}$, where 
    ${o_i \in O}$ is an object, and ${(x_i, y_i)}$ is the centroid of 
    a \textit{target region} that $o_i$ needs to be pushed into. The target
    region has a constant diameter $d$. When ${o_i = o_g}$, this is interpreted
    as the high-level action to reach for the goal object, and the centroid can
    be ignored.
    The high-level actions may be suggested by an automated high-level planner
    or by a human-operator.

    Consider \cref{fig:guidance_explained} as an example.
    A human-operator suggests a high-level action ${(o_2, 1.15, 0.4)}$
    (\cref{fig:guidance_explained_1}), where $(1.15, 0.4)$ is the centroid of the
    blue target region. The Guided-RTC system finds the controls to push
    $o_2$ into the target region (\cref{fig:guidance_explained_3}). When
    the human-operator suggests to reach for the goal object $o_g$
    (\cref{fig:guidance_explained_4}), the system finds the controls to perform 
    this action (\cref{fig:guidance_explained_5,fig:guidance_explained_6}). 

    In this work, we investigate how a Guided-RTC system
    with a human-in-the-loop performs when compared with (a) solving the original RTC problem
    directly using kinodynamic approaches (\cref{sec:rtc_planners}), and (b)
    using Guided-RTC systems with automated ways of generating the high-level 
    actions.

    In \cref{sub:grtc_planner} we present a generic algorithm to implement the
    Guided-RTC framework which is agnostic to how the high-level actions are
    generated. Then we present different approaches to generate the high-level 
    actions, including a human-in-the-loop approach
    in \cref{sub:grtc_hitl_planner}, as well as two other automated approaches in
    \cref{sub:namo_planner,sub:grtc_heuristic_planner}.

    \subsection{A Generic approach for Guided-RTC Planning}%
        \label{sub:grtc_planner}

        We present a generic algorithm for Guided-RTC in \cref{alg:grtc}. 
        The initial configuration of the problem is assumed to be the current
        configuration, $q_{current}$, of the system (\cref{algline:qcurrent}).
        The next high-level action is decided based on the current
        configuration (\cref{algline:get_next_action}).  If the object in the
        high-level action is not the goal object (\cref{algline:ifnotgoal}),
        then it is pushed to the target region between
        \cref{algline:approaching_states,algline:first_execution},
        and a new high-level action is requested.
        If it is the goal object, the robot tries to reach it between 
        \cref{algline:reach_goal,algline:second_execution} and the
        system terminates.

\begin{figure}[!t]
            \vspace*{2mm}
            \tiny
            \centering
            \def\svgwidth{0.9\columnwidth}
            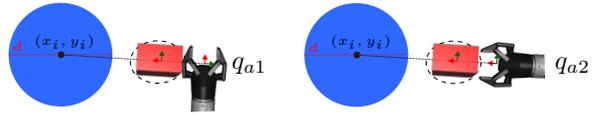
            \caption{Approaching states: The blue circle is the target region, the red rectangle the object to
            manipulate. We compute two approaching states, $q_{a1}$ and $q_{a2}$.}
            \label{fig:approaching_explained}
        \end{figure}

        We plan to push an object to its target region in two steps. In
        \cref{algline:first_kinodynamic_planning} we plan to an intermediate
        \textit{approaching state} near the object, and then in
        \cref{algline:second_kinodynamic_planning}, we plan from this
        approaching state to push the object to its target region.
        Specifically, given an object to push, $o_i$, we compute two
        approaching states $q_{a1}$ and $q_{a2}$
        (\cref{algline:approaching_states}).
        \cref{fig:approaching_explained} shows how these approaching
        states are computed, based on the object's current
        position, the centroid $(x_i, y_i)$ and the minimum enclosing circle
        of the object. The approaching state $q_{a1}$ encourages side-ways
        pushing, where $q_{a2}$ encourages forward pushing. 
        We also experimented with planning
        without first approaching the object but we found that
        approaching the object from a good pose yields to faster pushing 
        solutions.
        Using both approaching states as the goal we plan to move to one of them
        (multi-goal planning) in \cref{algline:first_kinodynamic_planning}.
        Then, from the approaching state reached (either $q_{a1}$ or $q_{a2}$) we
        push $o_i$ to its target region (\cref{algline:second_kinodynamic_planning}).
        If any of the two planning calls in \cref{algline:first_kinodynamic_planning,algline:second_kinodynamic_planning} fails, then the algorithm proceeds to the
        next high-level action (\cref{algline:get_next_action}). Otherwise,
        we execute the solutions sequentially in
        \cref{algline:first_execution}, which changes the current system configuration $q_{current}$.

        %Once the high-level action on \cref{algline:get_next_action} 
        %indicates $o_i = o_g$ we plan to reach for the goal object on
        %\cref{algline:reach_goal}.
        We use kinodynamic planners (e.g. kinodynamic RRT or KPIECE)
        to support the planning in
        \cref{algline:first_kinodynamic_planning,algline:second_kinodynamic_planning,algline:reach_goal}. 

        \begin{algorithm}[!b]
			\caption{Guided-RTC}\label{alg:grtc}
			\algrenewcommand\algorithmicindent{0.8em}%
			\begin{algorithmic}[1]
				\Procedure{GRTC}{$\stateSpace, \controlSpace, \initialState, \goalStates$}
                \State $q_{current} \gets q_0$ \label{algline:qcurrent}
					\Do
                        \State $o_i, x_i, y_i \gets$ \textproc{NextHighLevelAction($q_{current}$)} \label{algline:get_next_action}
                        \If{$o_i \neq o_g$} \label{algline:ifnotgoal}
                            \State $q_{a1}, q_{a2} \gets$ compute approaching states to $o_i$ \label{algline:approaching_states}
                            \State kinodynamicPlanning($q_{current}, \{q_{a1}, q_{a2}\}$) \label{algline:first_kinodynamic_planning}

                            \State \textbf{if} planning fails \textbf{then} \textbf{continue}
							%\If{planning fails}
                                %\State \textbf{continue}
							%\EndIf

                            \State kinodynamicPlanning($q_{a1}$ or $q_{a2}, (o_i, {x_i, y_i})$) \label{algline:second_kinodynamic_planning}
                            \State \textbf{if} planning fails \textbf{then} \textbf{continue}
							%\If{planning fails}
                                %\State \textbf{continue}
							%\EndIf
                            \State $q_{current} \gets$ execute solutions from \cref{algline:first_kinodynamic_planning,algline:second_kinodynamic_planning} \label{algline:first_execution}
						\EndIf
					\doWhile{$o_i \neq o_g$}
                    \State kinodynamicPlanning($q_{current}, \goalStates$) \label{algline:reach_goal}
					\If{planning succeeds}
                        \State $q_{current} \gets$ execute solution from \cref{algline:reach_goal} \label{algline:second_execution}
					\EndIf
				\EndProcedure
			\end{algorithmic}
		\end{algorithm}

        \cref{alg:grtc} runs up to an overall time limit, $T_{overall}$, or
        until a goal is reached. The pushing planning calls in
        \cref{algline:first_kinodynamic_planning,algline:second_kinodynamic_planning}
        have their own shorter time limit, $T_{pushing}$, and they should find a
        valid solution within this limit.
        %the algorithm goes back to
        %\cref{algline:get_next_action} for a new high level action. 
        The planning call in \cref{algline:reach_goal} is allowed to
        run until the overall time limit is over.

    \subsection{\ourplannerinfullacronym}%
        \label{sub:grtc_hitl_planner}

        \begin{algorithm}[!b]
			\caption{\ourplanneracronym}\label{alg:grtc_hitl}
			\algrenewcommand\algorithmicindent{0.8em}%
			\begin{algorithmic}[1]
                \Function{NextHighLevelAction}{$q_{current}$}
					\State $o_i \gets$ get object selection from human operator
					\If{$o_i \neq o_g$}
						\State $x_i, y_i \gets$ get region centroid from human operator
						\State \textbf{return} $o_i, x_i, y_i$
					\EndIf
					\State \textbf{return} $o_g$
				\EndFunction
			\end{algorithmic}
		\end{algorithm}

        \ourplannerinfullacronym is an instantiation of the \acronymourapproach
        Framework. A human-operator, through a graphical user interface, provides the 
		high-level actions. In \cref{alg:grtc_hitl} we present
        \ourplanneracronym \textproc{NextHighLevelAction} function 
        (referenced in \cref{alg:grtc}, \cref{algline:get_next_action}).

		The human provides high-level actions until she selects the goal 
		object, $o_g$. The \acronymourapproach framework (\cref{alg:grtc}) 
		plans and executes them. The state of the system changes after each 
		high-level action and the human operator is presented with the resulting state 
        each time ($q_{current}$). Note here that \textit{the operator can decide not to
        provide any guidance} (by selecting the goal object straightaway),
        which would be equivalent to running a state-of-the-art
        kinodynamic planning on the original RTC problem.

        We developed a simple user interface to 
		communicate with the human-operator. The operator at every step 
		is presented with a window showing the environment and the robot. 
		The operator, using a mouse pointer, provides the input by first clicking 
		on the desired object and then a point on the 
		plane (\cref{fig:guidance_explained_1}) that becomes the 
		centroid of the target region.

        The approach we propose here uses a human-operator to decide on the
        high-level plan. One question is whether one can use automatic
        approaches, and how they would perform compared to the human
        suggested actions. To make such a comparison, we implemented two automated 
        approaches.

    \subsection{Guided-RTC with NAMO} 
        \label{sub:namo_planner}
        We adapted the NAMO algorithm described by Stilman et al.
        \cite{stilman2007manipulation} to our problem as an alternative,
        autonomous, way to generate a high-level plan. NAMO has originally been used for
        pick-and-place manipulation. We adapted it to work for non-prehensile
        tasks by using a kinodynamic planner as the low-level planner, instead
        of collision-free motion planners as in the original work.

        To determine the ordering of objects to manipulate and where to place
        them, i.e. the high-level plan, NAMO uses backward planning. It starts by running
        the low-level planner to reach the goal, assuming the robot can travel through other movable
        objects. The resulting volume of space swept by the robot to
        reach the goal object is then checked to see which movable objects
        intersect with it. These objects are added to a queue to be moved out
        of this swept volume. The algorithm then pops out an object from this
        queue and makes a recursive call to reach and move that object. This process
        continues until the queue is empty, meaning that (1) there is a plan to
        reach and move every object out of the way, and (2) there is a position to
        place every object out of the accumulated robot swept volume.
        The last object planned for is the first one to be moved during execution. For a
        more detailed explanation of NAMO, we refer the reader to Stilman et
        al.  \cite{stilman2007manipulation}. 

        Since NAMO plans backward, to decide on the first object to be moved, it needs to determine all
        the objects to be moved and their target positions. While this means
        NAMO can offer theoretical guarantees when a plan exists, it also means
        that in highly cluttered environments like ours, NAMO can quickly run
        out of space to place objects, before it resolves all the constraints.
        In our experimental setting which includes a high number of objects in a
        restricted shelf space, NAMO failed in all cases by filling up the space with
        the robot swept volume before a plan for all objects in the queue have been found.

        This motivated us to design a heuristic approach similar to NAMO, but
        one that plans forward, by directly identifying the first object to move out of
        the way.

    \subsection{Guided-RTC with Straight Line Heuristic (\heuristicplannername)}%
        \label{sub:grtc_heuristic_planner}

        \begin{figure}[!t]
            \vspace*{2mm}
            \captionsetup[subfigure]{aboveskip=1.0pt,belowskip=1.0pt}
            \centering
            \begin{subfigure}{.33\linewidth}
                \centering
                \tiny
                \def\svgwidth{0.9\columnwidth}
                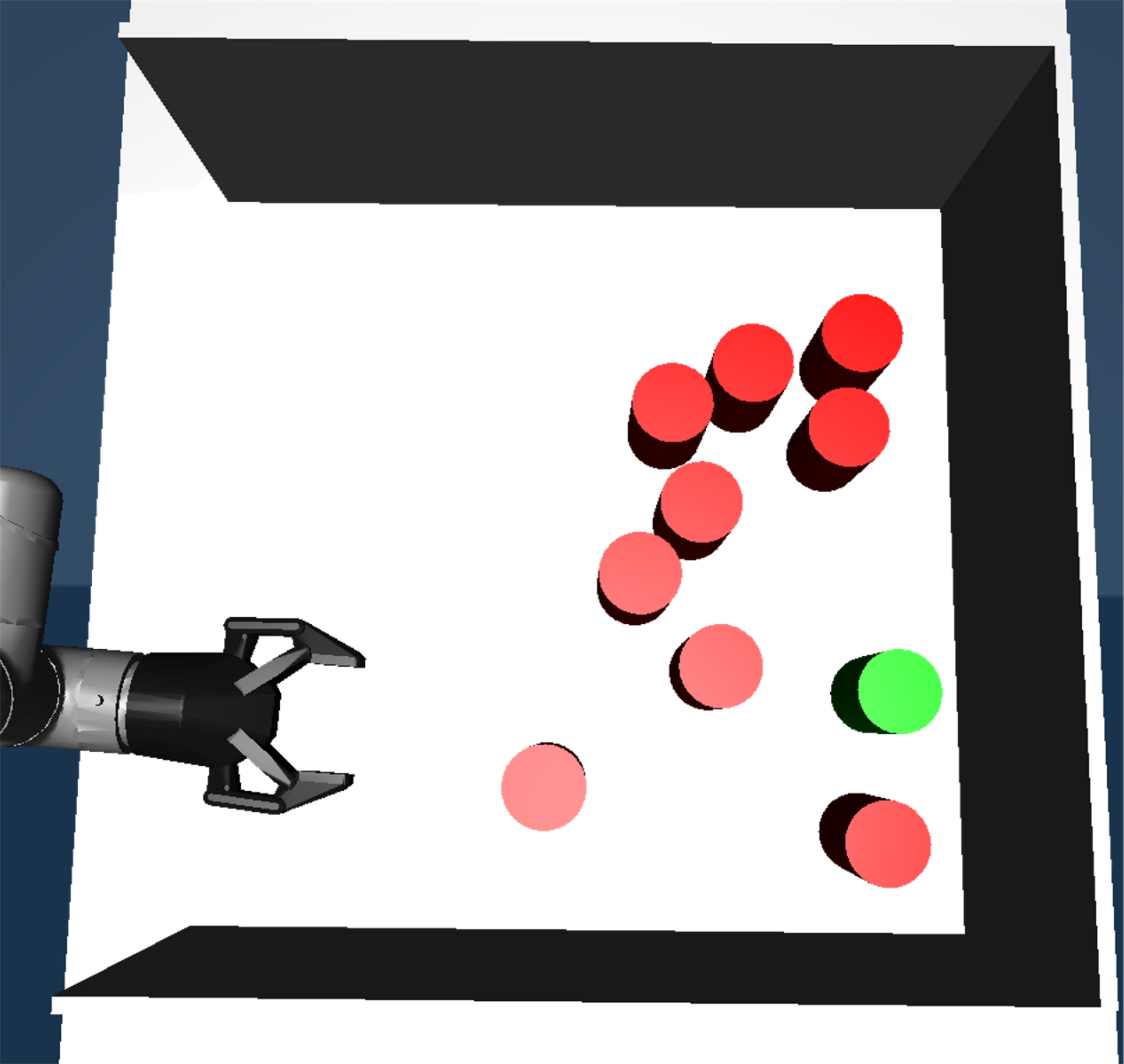
                \caption{}
                \label{fig:heuristic_explained_1}
            \end{subfigure}%
            \begin{subfigure}{.33\linewidth}    
                \centering
                \tiny
                \def\svgwidth{0.9\columnwidth}
%% Accompanies image file '2.pdf' (pdf, eps, ps)
\begingroup%
  \makeatletter%
  \providecommand\color[2][]{%
    \errmessage{(Inkscape) Color is used for the text in Inkscape, but the package 'color.sty' is not loaded}%
    \renewcommand\color[2][]{}%
  }%
  \providecommand\transparent[1]{%
    \errmessage{(Inkscape) Transparency is used (non-zero) for the text in Inkscape, but the package 'transparent.sty' is not loaded}%
    \renewcommand\transparent[1]{}%
  }%
  \providecommand\rotatebox[2]{#2}%
  \newcommand*\fsize{\dimexpr\f@size pt\relax}%
  \newcommand*\lineheight[1]{\fontsize{\fsize}{#1\fsize}\selectfont}%
  \ifx\svgwidth\undefined%
    \setlength{\unitlength}{950.25bp}%
    \ifx\svgscale\undefined%
      \relax%
    \else%
      \setlength{\unitlength}{\unitlength * \real{\svgscale}}%
    \fi%
  \else%
    \setlength{\unitlength}{\svgwidth}%
  \fi%
  \global\let\svgwidth\undefined%
  \global\let\svgscale\undefined%
  \makeatother%
  \begin{picture}(1,0.94711918)%
    \lineheight{1}%
    \setlength\tabcolsep{0pt}%
    \put(0,0){\includegraphics[width=\unitlength,page=1]{2.pdf}}%
    \put(0.45086655,0.17464552){\color[rgb]{0,0,0}\makebox(0,0)[lt]{\lineheight{1.25}\smash{\begin{tabular}[t]{l}$o_7$\end{tabular}}}}%
  \end{picture}%
\endgroup%

                \caption{}
                \label{fig:heuristic_explained_2}
            \end{subfigure}%
            \begin{subfigure}{.33\linewidth}    
                \centering
                \tiny
                \def\svgwidth{0.9\columnwidth}
%% Accompanies image file '3.pdf' (pdf, eps, ps)
\begingroup%
  \makeatletter%
  \providecommand\color[2][]{%
    \errmessage{(Inkscape) Color is used for the text in Inkscape, but the package 'color.sty' is not loaded}%
    \renewcommand\color[2][]{}%
  }%
  \providecommand\transparent[1]{%
    \errmessage{(Inkscape) Transparency is used (non-zero) for the text in Inkscape, but the package 'transparent.sty' is not loaded}%
    \renewcommand\transparent[1]{}%
  }%
  \providecommand\rotatebox[2]{#2}%
  \newcommand*\fsize{\dimexpr\f@size pt\relax}%
  \newcommand*\lineheight[1]{\fontsize{\fsize}{#1\fsize}\selectfont}%
  \ifx\svgwidth\undefined%
    \setlength{\unitlength}{950.25bp}%
    \ifx\svgscale\undefined%
      \relax%
    \else%
      \setlength{\unitlength}{\unitlength * \real{\svgscale}}%
    \fi%
  \else%
    \setlength{\unitlength}{\svgwidth}%
  \fi%
  \global\let\svgwidth\undefined%
  \global\let\svgscale\undefined%
  \makeatother%
  \begin{picture}(1,0.94711918)%
    \lineheight{1}%
    \setlength\tabcolsep{0pt}%
    \put(0,0){\includegraphics[width=\unitlength,page=1]{3.pdf}}%
    \put(0.52515689,0.26180397){\color[rgb]{0.48627451,0.44313725,0}\rotatebox{-0.06066708}{\makebox(0,0)[lt]{\lineheight{1.25}\smash{\begin{tabular}[t]{l}$\mathit{V_{swept}}$\end{tabular}}}}}%
  \end{picture}%
\endgroup%

                \caption{}
                \label{fig:heuristic_explained_3}
            \end{subfigure}
            \caption{\heuristicplannername: (a) Initial state. 
                (b) The robot moves on a straight line to the goal object, $o_g$, to
                obtain the first blocking obstacle ($o_7$) and the swept volume
                (yellow area). (c) The heuristic produces a high-level action
                for $o_7$ indicated by the arrow and the target region (blue).
                This process is repeated until $V_{swept}$ contains no 
                blocking obstacle.}
            \label{fig:heuristic_explained}
        \end{figure}

        \begin{algorithm}[!b]
            \caption{\heuristicplannername Planner}\label{alg:grtc_heuristic}
            \algrenewcommand\algorithmicindent{0.8em}%
            \begin{algorithmic}[1]
                \Function{NextHighLevelAction}{$q_{current}$}
                	\State $o_b \gets$ find the first blocking obstacle to $o_g$ \label{algline:find_first_blocking_obstacle}
					\If{there exists a blocking obstacle $o_b$}
                    	\State $x_b, y_b \gets$ {find collision-free placement of $o_b$} \label{algline:find_region}
                        \State \textbf{return} $o_b, x_b, y_b$ \label{algline:return_action}
					\EndIf
                    \State \textbf{return} $o_g$ \Comment{No blocking obstacle, reach the goal} \label{algline:return_goal}
                \EndFunction
            \end{algorithmic}
        \end{algorithm}
        
        We present this approach in \cref{alg:grtc_heuristic} and illustrate it in \cref{fig:heuristic_explained}. 
        This heuristic assumes the robot moves on a straight line from its current position towards the goal object (\cref{fig:heuristic_explained_2}).
        The first blocking object, $o_b$ in \cref{algline:find_first_blocking_obstacle}, is identified as the next object to be moved.
        During the straight line motion, we capture the robot's swept volume, $V_{swept}$
        (\cref{fig:heuristic_explained_2}). We randomly sample a
        collision-free target region centroid outside $V_{swept}$
        (\cref{alg:grtc_heuristic} \cref{algline:find_region} and
        \cref{fig:heuristic_explained_3}). The object and the centroid are then
        returned as the next high-level action (\cref{alg:grtc_heuristic}
        \cref{algline:return_action}). The centroid sampling happens 30cm around 
        the object's initial position to maximize the chance of a successful
        pushing. If there is no collision-free space around $o_b$, 
        then we sample from the entire space.

        After every high-level action suggested by the heuristic, the Guided-RTC framework
        (\cref{alg:grtc}) plans and executes it and the state of the system is updated
        ($q_{current}$). The heuristic then suggests a new high-level action
        from $q_{current}$ until there is no blocking obstacle (\cref{alg:grtc_heuristic} \cref{algline:return_goal}).

    \section{Experiments \& Results}
    \label{sec:experiment_results}

    The algorithms we evaluate are: (1) \ourplanneracronym which is the main
    algorithm we propose and uses a human operator to obtain the high-level
    actions (\cref{sub:grtc_hitl_planner}), (2) \heuristicplannername which uses a
    straight line heuristic to automatically determine the high-level actions
    (\cref{sub:grtc_heuristic_planner}) and (3) Kinodynamic RRT and KPIECE
    (\cref{sec:rtc_planners}) which plan to reach for the goal object
    without a high-level plan. As explained in \cref{sub:namo_planner}, NAMO failed to find solutions in our
    problems and therefore we did not include results for it here. 

    For all experiments, we use the Open Motion Planning Library (OMPL)
    \cite{omplSucan2012} implementation of RRT and KPIECE.  We use MuJoCo\footnote{On 
    a computer with Intel Core i7-4790 CPU @ 3.60GHz, 16GB RAM.}
    \cite{mujocoTodorov2012} to implement the system dynamics, $f$.  
    For all planners, the overall planning time limit, $T_{overall}$, is 300
    seconds, after which it was considered a failure. 
    For \ourplanneracronym and \heuristicplannername, $T_{pushing}$ is 10 seconds.
    For \ourplanneracronym, the human-interaction time was included in the
    overall time limit.  The same human-operator, who was experienced with the system,
    was used in all experiments.  Since we are interested in an
    industrial/warehouse scenario where human-operators would be trained to use the
    system, we are mainly interested in the performance of trained operators,
    rather than novices. 

    In \cref{sub:simulation_results} we present simulation results comparing
    \ourplanneracronym  with Kinodynamic RRT, KPIECE, and \heuristicplannername.
    In \cref{sub:parallel_guidance} we show results where the human operator guides multiple robots in parallel, in simulation.
    In \cref{sub:real_robot_results} we present real-world experiments comparing 
    kinodynamic planners with \ourplanneracronym on 10 different scenes.
    A video with some of these experiments is available on \url{https://youtu.be/nfr1Fdketrc}.

    \begin{figure}[!t]
        \vspace*{2mm}
        \captionsetup[subfigure]{aboveskip=1.0pt,belowskip=1.0pt}
        \centering
        \begin{subfigure}{.24\linewidth}
            \centering
            \def\svgwidth{0.95\columnwidth}
            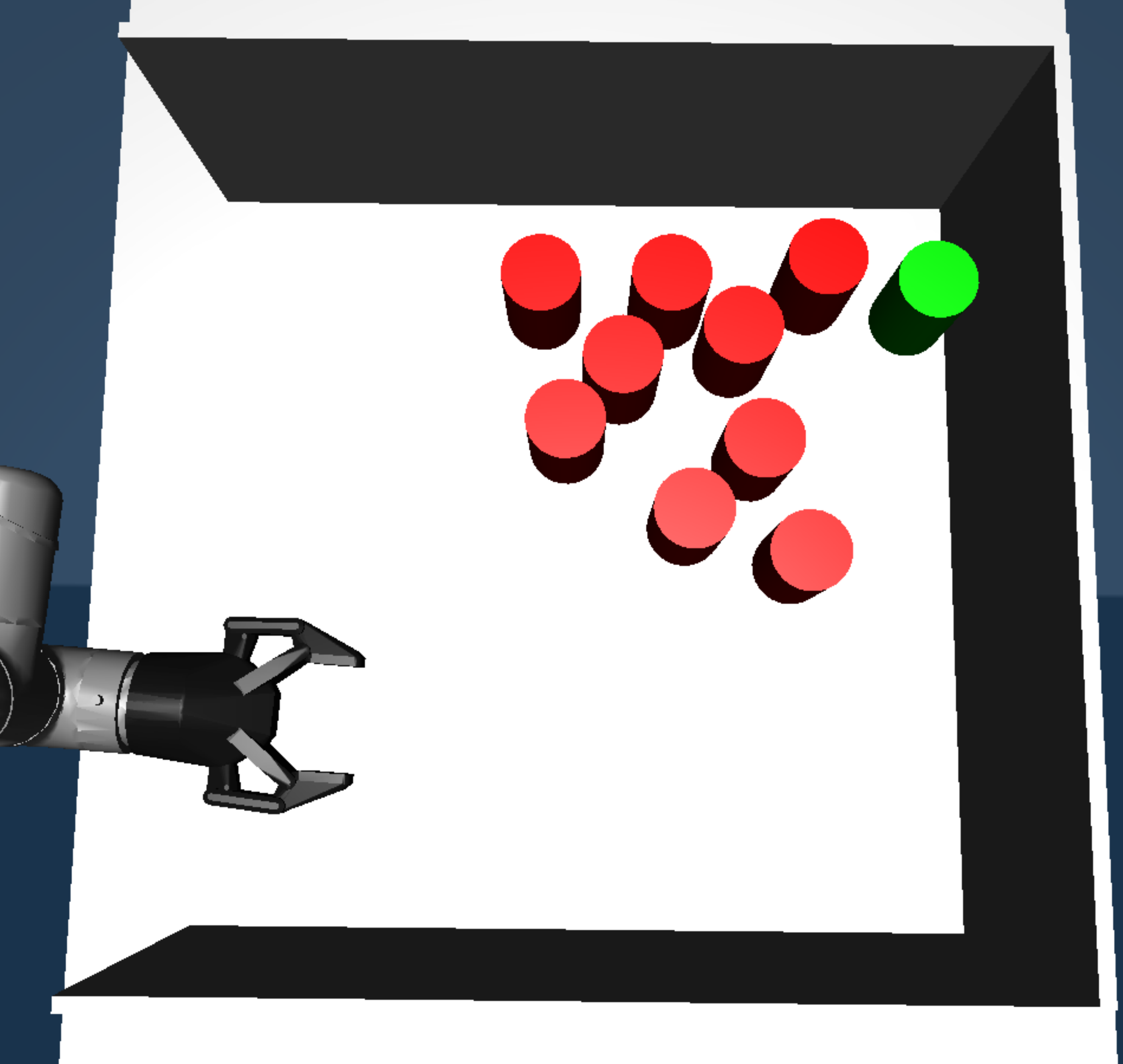
            \caption{S1}
            \label{fig:p1}
        \end{subfigure}%
        \begin{subfigure}{.24\linewidth}
            \centering
            \def\svgwidth{0.95\columnwidth}
            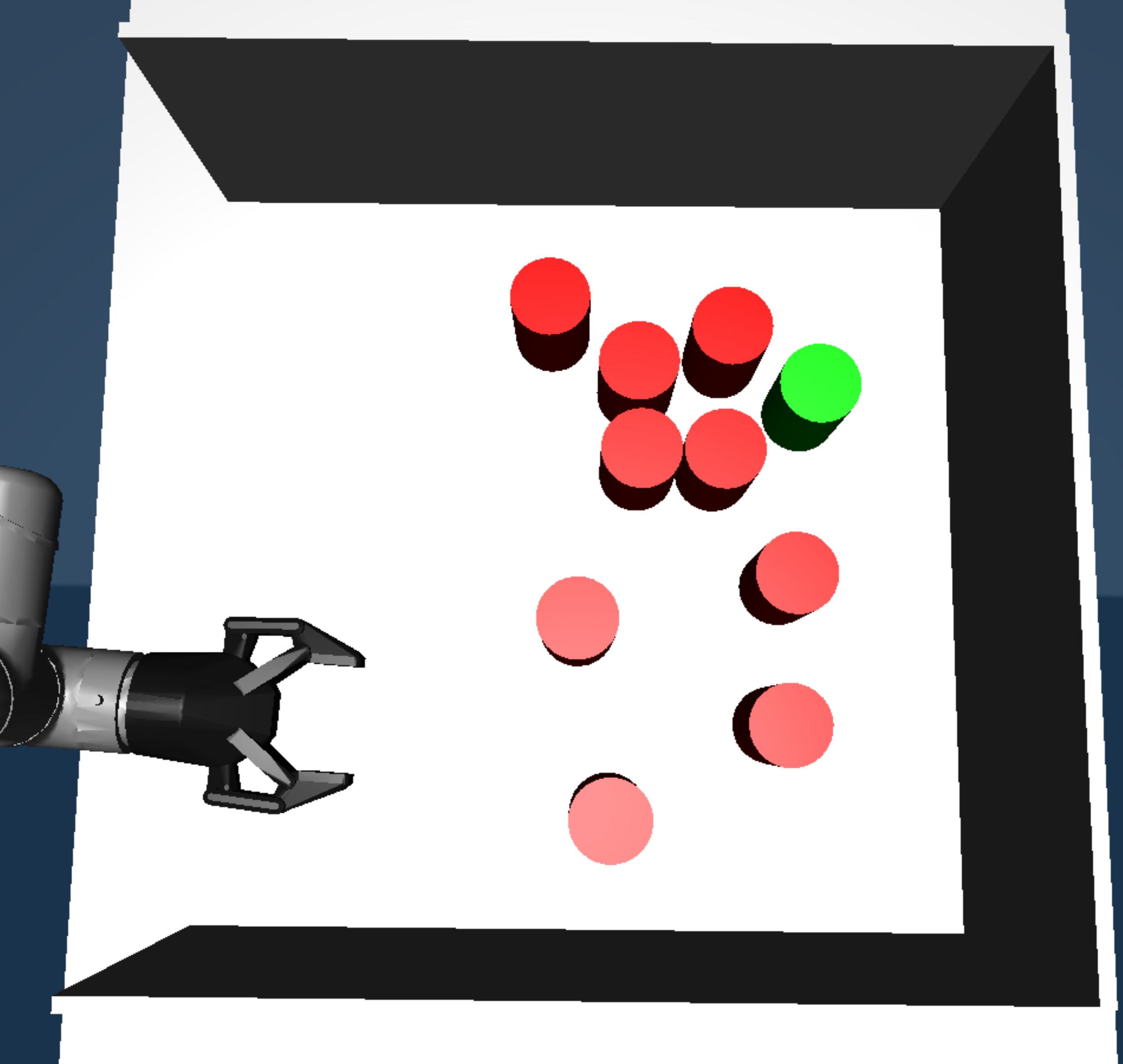
            \caption{S2}
            \label{fig:p2}
        \end{subfigure}%
        \begin{subfigure}{.24\linewidth}
            \centering
            \def\svgwidth{0.95\columnwidth}
            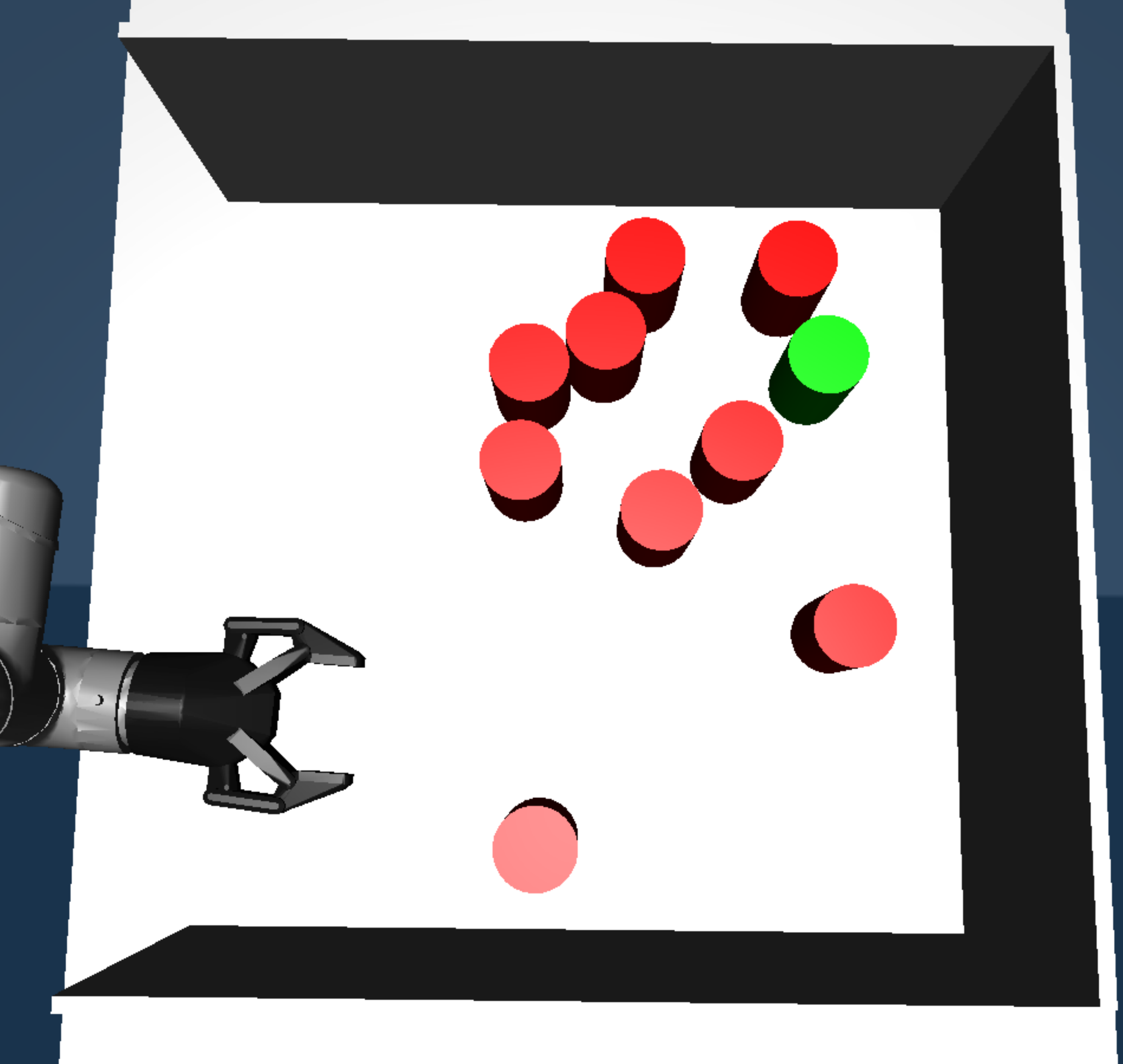
            \caption{S3}
            \label{fig:p3}
        \end{subfigure}%
        \begin{subfigure}{.24\linewidth}
            \centering
            \def\svgwidth{0.95\columnwidth}
            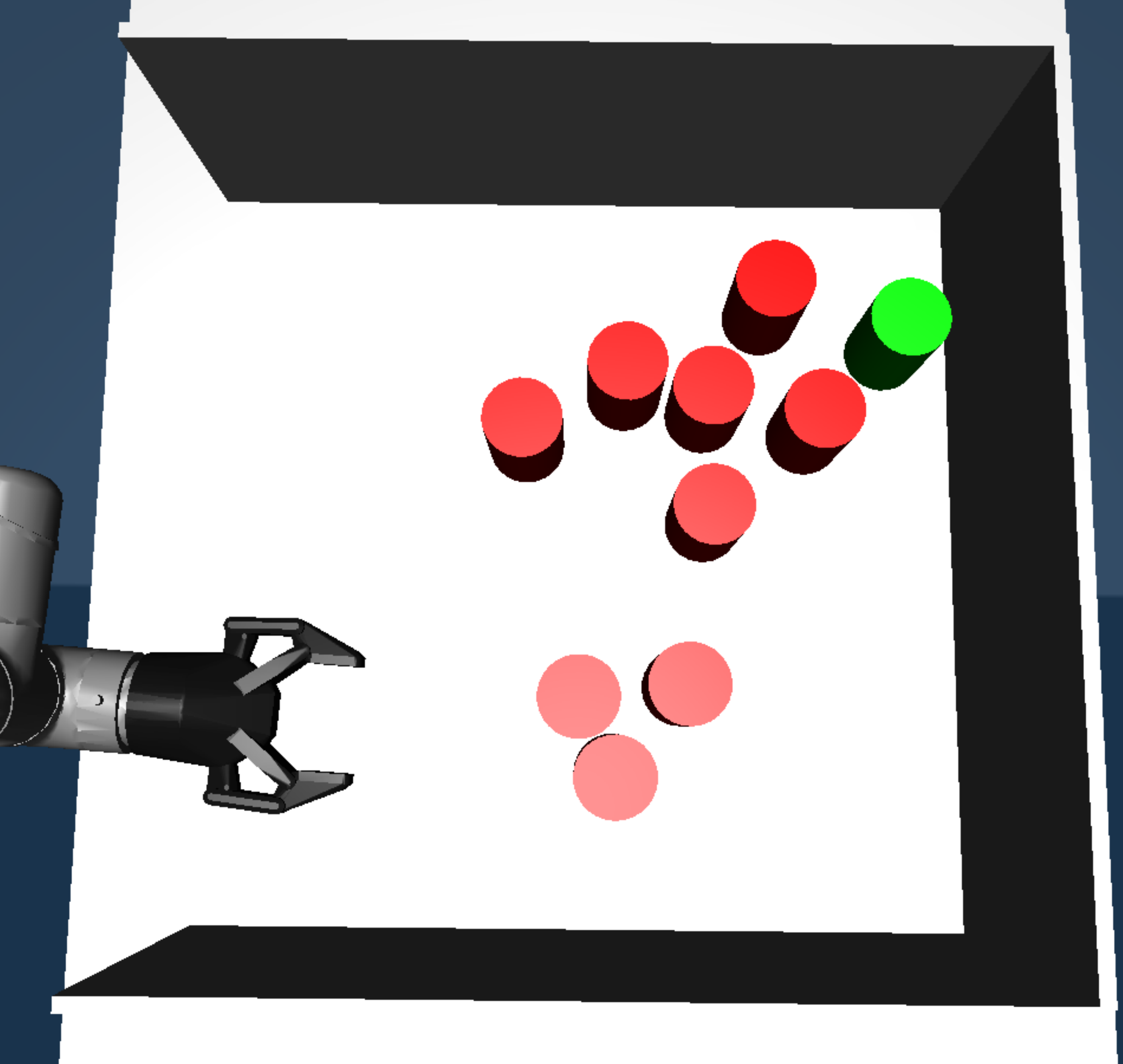
            \caption{S4}
            \label{fig:p4}
        \end{subfigure}

        \begin{subfigure}{.24\linewidth}
            \centering
            \includegraphics[width=0.95\linewidth]{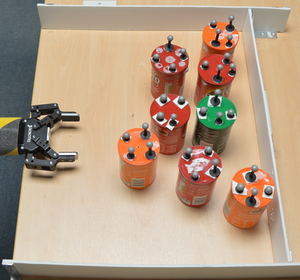}
            \caption{R1}
            \label{fig:r1}
        \end{subfigure}%
        \begin{subfigure}{.24\linewidth}
            \centering
            \includegraphics[width=0.95\linewidth]{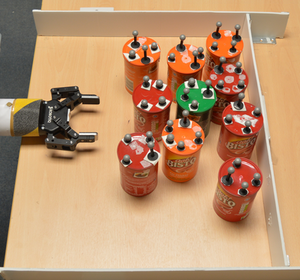}
            \caption{R2}
            \label{fig:r2}
        \end{subfigure}%
        \begin{subfigure}{.24\linewidth}
            \centering
            \includegraphics[width=0.95\linewidth]{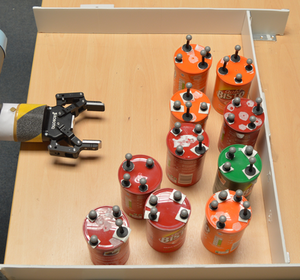}
            \caption{R3}
            \label{fig:r3}
        \end{subfigure}%
        \begin{subfigure}{.24\linewidth}
            \centering
            \includegraphics[width=0.95\linewidth]{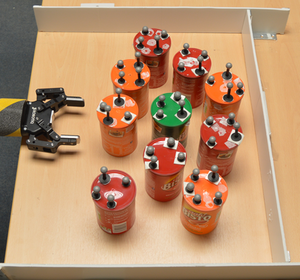}
            \caption{R4}
            \label{fig:r4}
        \end{subfigure}
        \caption{Initial states of different problems in simulation
        (S1-S4) and real world (R1-R4). Goal object is in green.
        %($o_g$), red and orange objects are movable obstacles ($O$).
        }
        \label{fig:scenes}
    \end{figure}

    \subsection{Simulation Results}%
        \label{sub:simulation_results}

        \begin{figure}[!t]
            \vspace*{2mm}
            \centering

            \scriptsize
            \def\svgwidth{\columnwidth}
            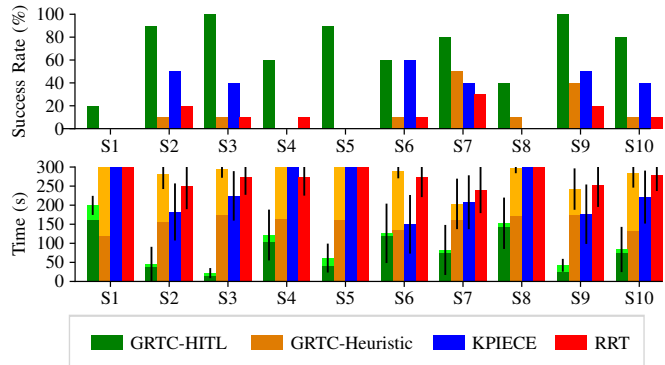

            \caption{Simulation results, for each scene (S1-S10): (Top) Success 
            rate. (Bottom) Mean planning time. The error bars indicate 
            the 95\% CI. For \ourplanneracronym and \heuristicplannername, the 
            dark shade indicates the planning time where the light shade 
            indicates the time it took to produce the high-level actions (for 
            \ourplanneracronym this is a fraction of the time).}
            \label{fig:simulation_results}
        \end{figure}

        We evaluated each approach 100 times by running them 10 times in 10
        different, randomly-generated, scenes. We use a randomizer that places the goal 
        object at the back of the shelf and then incrementally places the 
        remaining (nine) objects in the shelf such that no object collides with 
        each other. Some of the scenes are presented in \crefrange{fig:p1}{fig:p4}.

        For \ourplanneracronym, the human-operator interacted with each scene 
        \textit{once} and from the last state left by the human-operator we ran 
        the planner (\cref{alg:grtc} \cref{algline:reach_goal}) to reach for
        the goal object \textit{10 times}.
        For \heuristicplannername we ran all 100 experiments with both RRT and
        KPIECE as the low-level planners and we present the better performing
        one. For \ourplanneracronym and \heuristicplannername the low-level 
        planner is RRT.

        \cref{fig:simulation_results} summarizes the results of our experiments
        for each of the random scenes (S1-S10).
        \cref{fig:simulation_results}-Top shows that \ourplanneracronym yields
        to more successes per scene than any other approach except for S6 which
        was as successful as KPIECE. The overall success 
        rate for each approach is 72\% for \ourplanneracronym, 11\% for RRT, 28\% for 
        KPIECE and 14\% for \heuristicplannername.
        \cref{fig:simulation_results}-Bottom shows that
        \ourplanneracronym improved the planning time in \textit{all} scenes. 

        \cref{tbl:interaction_results} summarizes the guidance performance for
        \ourplanneracronym and \heuristicplannername for all ten scenes. 
        Proposed Actions indicates the total number of high-level actions proposed. This number 
        includes the successful actions (actions that the planner managed to 
        satisfy) and failed actions (actions that the planner could not find 
        a solution for). Guidance Time indicates the time spent on generating 
        the high-level actions in seconds (in case of
        \ourplanneracronym the time the human-operator was interacting with the
        system and for \heuristicplannername the time took for the heuristic to
        generate the high-level actions). On average, the human proposed around 
        5 actions, of which around 3 were successful.
        On the other side, \heuristicplannername proposed on average 
        around 88 actions, of which only 3 were successful. The human 
        operator spent on average 14 seconds interacting with the system 
        while \heuristicplannername spent on average 124 seconds proposing
        high-level actions.

    \subsection{Parallel Guidance}%
        \label{sub:parallel_guidance}

        \begin{table}[!b]
          \centering
          \caption{Simulation results.}
          \label{tbl:interaction_results}
          \begin{tabular}{|l|cc|cc|}
                    \hline
                    \multirow{2}{*}{} &
                    \multicolumn{2}{c|}{\textbf{\ourplanneracronym}} &
                    \multicolumn{2}{c|}{\textbf{\heuristicplannername}} \\
                    \cline{2-5}
                    & $\mu$ & $\sigma$ & $\mu$ & $\sigma$\\
                    \hline
                    Proposed Actions & 4.9 & 3.3 & 88.4 & 58.2 \\
                    Successful Actions & 3.1 & 1.0 & 3.0 & 1.4 \\
                    Guidance Time (s) & 13.6 & 10.0 & 124.3 & 81.7 \\
                    \hline
                  \end{tabular}
        \end{table}%

        \begin{table}[!b]
          \centering
          \caption{Parallel vs. Individual Guidance.}
          \label{tbl:simultaneous}
          \begin{tabular}{|l|cc|cc|}
            \hline
            \multirow{2}{*}{} & \multicolumn{2}{c|}{\textbf{Parallel}} & \multicolumn{2}{c|}{\textbf{Individual}} \\
            \cline{2-5}
            & $\mu$ & $\sigma$ & $\mu$ & $\sigma$\\
            \hline
            Guidance Time (s) & 21.1 & 28.0 & 13.0 & 14.4 \\
            Overall Planning Time (s) & 139.7 & 117.9 & 122.8 & 120.0 \\
            Planner Idle Time (s) & 14.8 & 9.4 & 0.0 & 0.0 \\
            \hline
          \end{tabular}
        \end{table}

        \begin{table}[!b]
          \centering
          \caption{Real-world results.}
          \label{tbl:real_robot_results}
          \begin{tabular}{|l|c|c|c|}
            \hline
            \multirow{1}{1.47cm}{} &
            \multicolumn{1}{c|}{\textbf{\ourplanneracronym}} &
            \multicolumn{1}{c|}{\textbf{KPIECE}} &
            \multicolumn{1}{c|}{\textbf{RRT}}\\
            \hline
            Successes & 7 & 1 & 2\\
            Planning Failures & 2 & 4 & 8\\
            Execution Failures & 1 & 5 & 0\\ \hline
          \end{tabular}
        \end{table}

\begin{figure}[!t]
            \vspace*{2mm}
            \centering
            \fontsize{5pt}{5pt}\selectfont
            \def\svgwidth{\columnwidth}
            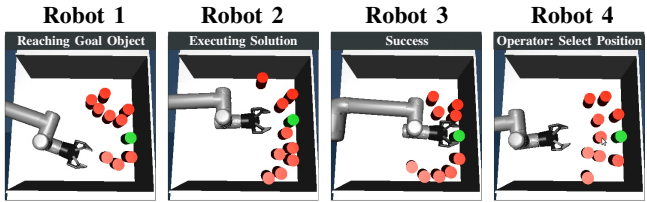
            \caption{Parallel Guidance: The first robot is planning to reach for 
            the goal object, the second one executes a solution, the third robot 
            successfully reached the goal object, the fourth robot is waiting 
            for human input (operator's main focus).}
            \label{fig:parallel_guidance}
        \end{figure}

        Since the human spends only a small amount of time guiding the robot using \ourplanneracronym, a single human-operator can be 
        used to guide multiple robots simultaneously, e.g. in a warehouse where a small number of operators
        remotely guide a large number of robots. We present an example in \cref{fig:parallel_guidance}.
        We performed experiments to test how the performance was affected when 
        the human-operator guides multiple robots in parallel. We tested guiding four robots in parallel. We generated 20 new
        random scenes and divided them into five groups of four. We compare the
        performance when the robots were guided individually in 20 different
        runs, to the performance when four robots were guided in parallel in
        five different runs. We repeated this process two times (i.e. 40 individual
        runs and 10 parallel runs). 
        Note that in the worst case with a time limit of 300 seconds, running
        20 scenes individually requires 100 minutes of human-operator time;
        whereas five parallel runs require just 25 minutes.

        \cref{tbl:simultaneous} summarizes these results. The success
        rate of parallel guidance is 60\% and for individual guidance 70\%.
        This efficient use of the human-operator's time comes with the cost of 
        slightly increased planning time and lower success rate. We also measured 
        the time the system was waiting for human input which was on average 
        15 seconds.

    \subsection{Real-robot results}%
        \label{sub:real_robot_results}

        We performed experiments using a UR5 manipulator on a Ridgeback
        omnidirectional base. We used the OptiTrack motion capture system to detect
        initial object/robot poses and to update the state in the human interface after
        every high-level action.

        We evaluated RRT, KPIECE and \ourplanneracronym performance in \textit{ten} different
        problems in the real world. We show some of the scenes 
        in \crefrange{fig:r1}{fig:r4}. 

        \begin{figure}[!t]
            \vspace*{2mm}
            \captionsetup[subfigure]{aboveskip=1.0pt,belowskip=1.0pt}
            \centering
            \begin{subfigure}{.25\linewidth}
                \centering
                \def\svgwidth{0.96\columnwidth}
                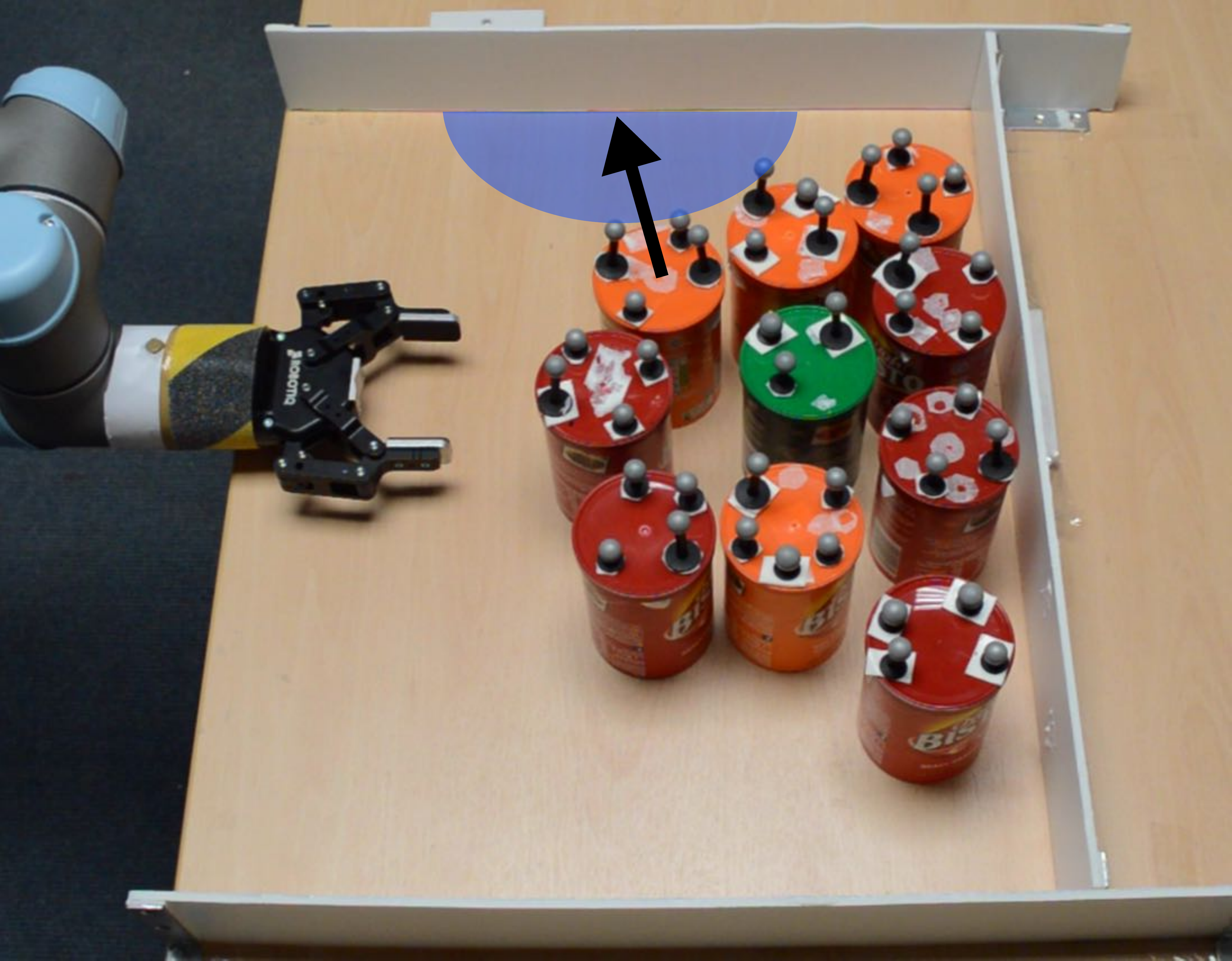
                \caption{}
                \label{fig:real_experiment_explained_2_1}
            \end{subfigure}%
            \begin{subfigure}{.25\linewidth}
                \centering
                \def\svgwidth{0.96\columnwidth}
                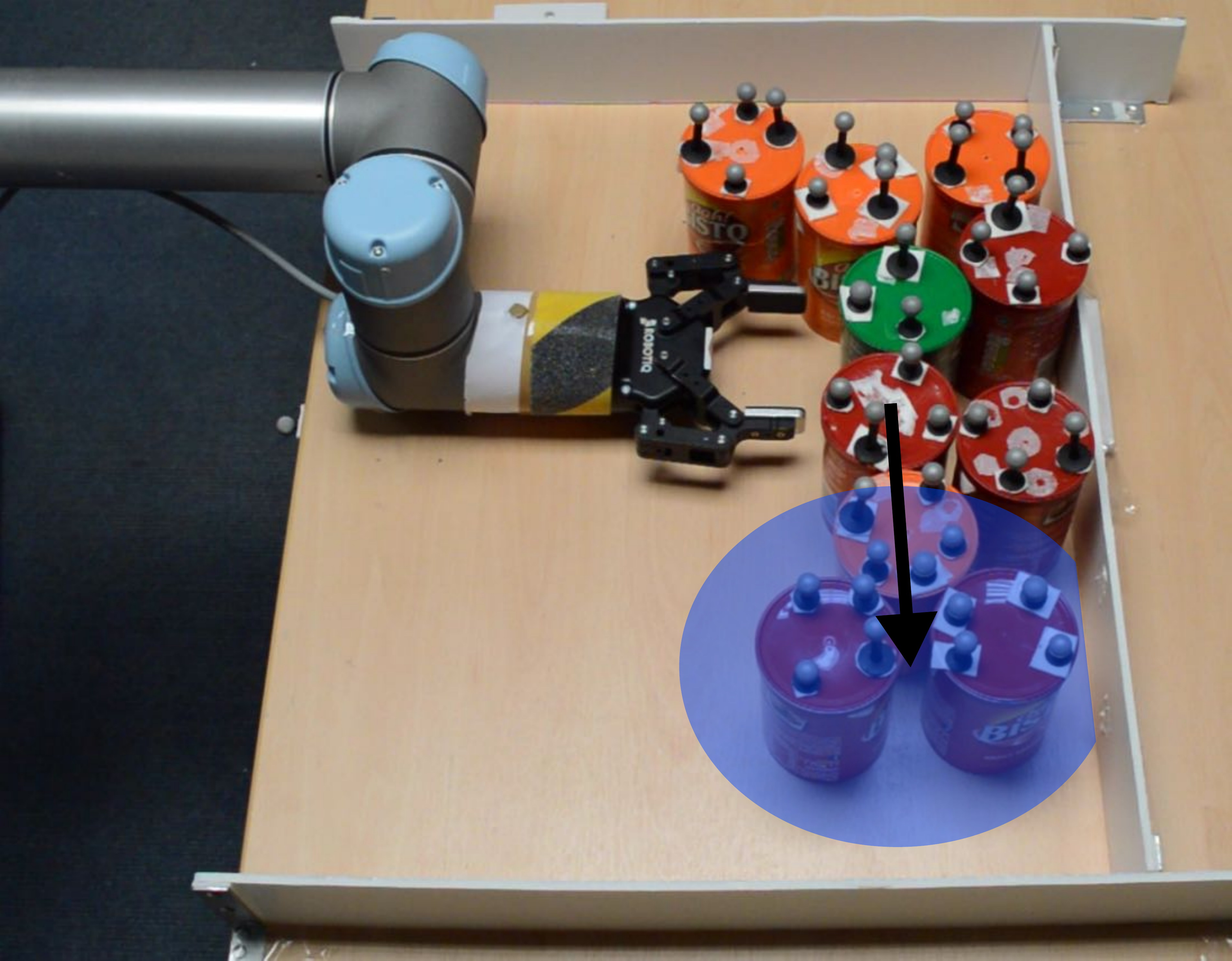
                \caption{}
                \label{fig:real_experiment_explained_2_2}
            \end{subfigure}%
            \begin{subfigure}{.25\linewidth}
                \centering
                \def\svgwidth{0.96\columnwidth}
                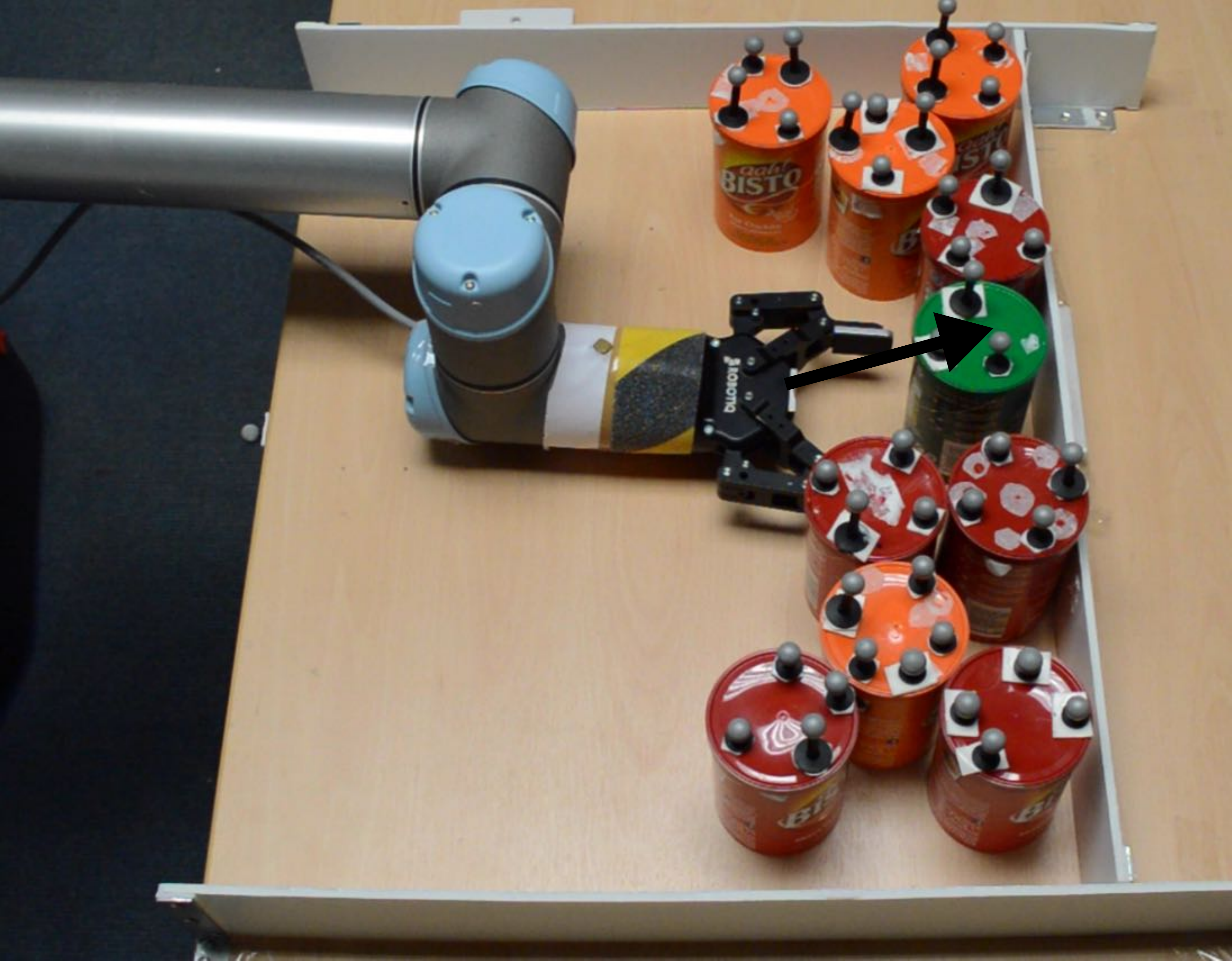
                \caption{}
                \label{fig:real_experiment_explained_2_3}
            \end{subfigure}%
            \begin{subfigure}{.25\linewidth}
                \centering
                \def\svgwidth{0.96\columnwidth}
                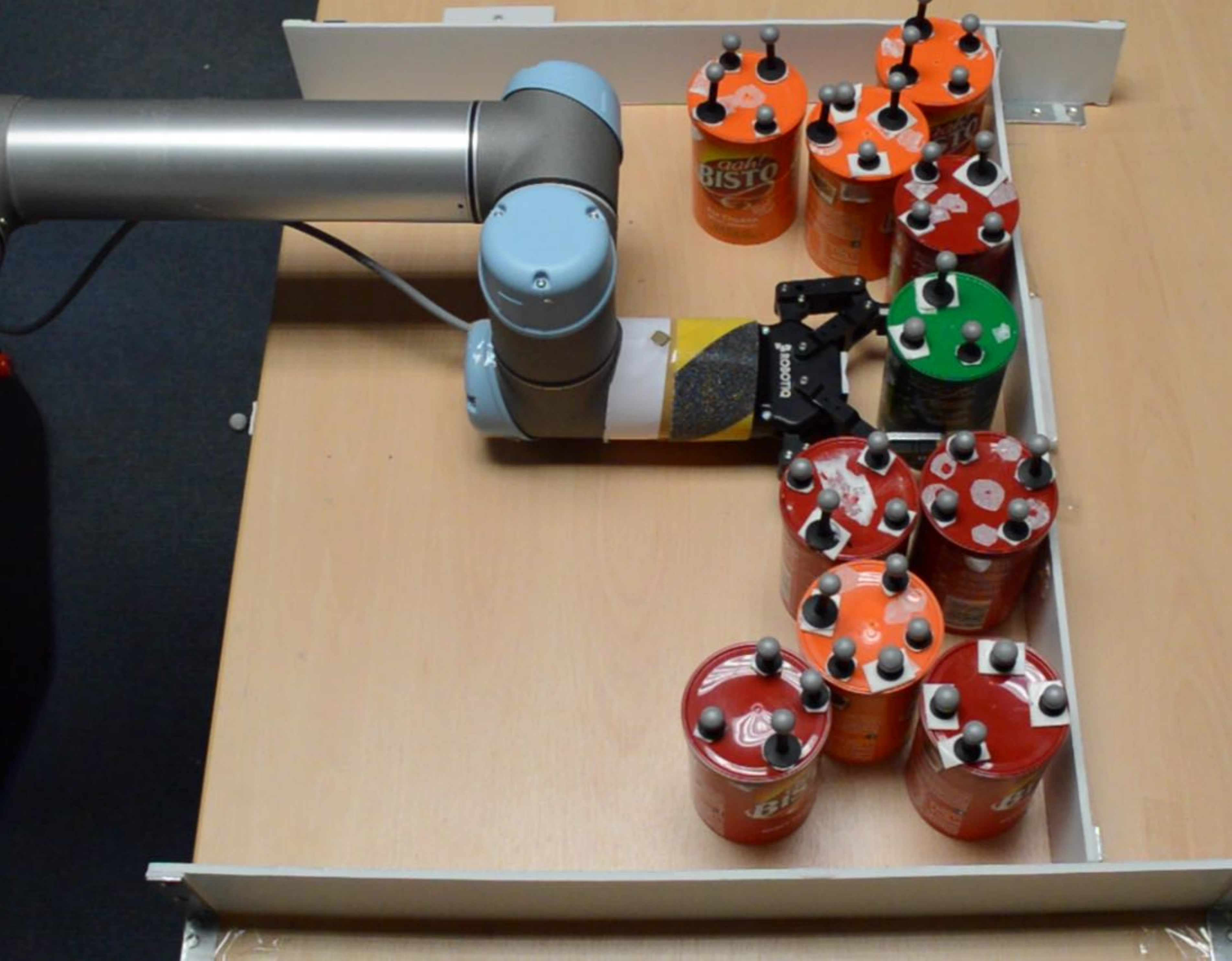
                \caption{}
                \label{fig:real_experiment_explained_2_4}
            \end{subfigure}
            \caption{A guided planning demonstration in the real world.}
            \label{fig:real_experiment_explained_2}
        \end{figure}

        \cref{tbl:real_robot_results} summarizes the success rate of each
        approach in the real world. When we say that the robot failed during execution,
        we mean that although the planner found a solution,
        when the robot executed the solution in the real-world, it 
        either failed to reach the goal 
        object, or it violated some constraint
        (hit the shelf or dropped an object to the floor). 
        These execution failures were due to the uncertainty in the real world:
        The result of the robot's 
        actions in the real-world yield to different states than the ones 
        predicted by the planner.

        The success rate for \ourplanneracronym, RRT and KPIECE is 70\%, 20\%,
        and 10\% respectively. \ourplanneracronym failed 20\% during planning
        and 10\% during execution. KPIECE was more successful during planning
        than RRT but failed most of the times during execution. RRT, on the other,
        hand accounts for more failures during planning than any other approach.
        %Additionally, the 
        %state left by the human operator was usually less
        %cluttered than the initial state which minimized real-world failures when 
        %reaching for the goal object.

        In \cref{fig:real_experiment_explained_1,fig:real_experiment_explained_2}
        we show two examples. In the first example, the human operator
        provides the first high-level action in \cref{fig:real_experiment_explained_1_1}
        and then indicates the goal object in \cref{fig:real_experiment_explained_1_3} 
        which is reached in \cref{fig:real_experiment_explained_1_4}.
        In the second example, the human-operator provides initially two high-level actions
        (\cref{fig:real_experiment_explained_2_1} and
        \cref{fig:real_experiment_explained_2_2}). The operator in
        \cref{fig:real_experiment_explained_2_3} indicates 
        the goal object and the robot reached the goal object in
        \cref{fig:real_experiment_explained_2_4}.

    \section{Conclusions}
    \label{sec:conclusions}

    We introduced a new human-in-the-loop framework for physics-based
    non-prehensile manipulation in clutter (\ourplanneracronym). We showed through 
    simulation and real-world experiments that \ourplanneracronym is more
    successful and faster in finding solutions than the three baselines we
    compared with. We also presented experiments where a single
    human-operator guides multiple robots in parallel, to make best use of the
    operator's time. We made the source code of our framework and of the
    baselines publicly available.

    To the best of our knowledge, this is the first work to look into
    non-prehensile manipulation with human-in-the-loop. 
    In the future, we would like to build on this work to evaluate the parallel
    control of higher numbers of robots, by minimizing the time the system is
    idle.

    \addtolength{\textheight}{-2cm}

    % References
    \bibliographystyle{IEEEtran}
    \bibliography{refs}
\end{document}